\documentclass[sigconf]{acmart}
\usepackage{epstopdf}
\usepackage{amsmath}
\usepackage{cases}
\usepackage{color}
\usepackage{tabularx}
\usepackage{graphicx}
\usepackage{subfigure}
\usepackage{array}
\usepackage{hyperref}
\usepackage{makecell}
\usepackage{multirow}
\usepackage{subfigure}
\usepackage{colortbl}
\usepackage{textcomp}
\usepackage{pifont}
\AtBeginDocument{%
  }

\setcopyright{acmcopyright}
\copyrightyear{2023}
\acmYear{2023}
\setcopyright{acmlicensed}\acmConference[MM '23]{Proceedings of the 31st
	ACM International Conference on Multimedia}{October 29-November 3,
	2023}{Ottawa, ON, Canada}
\acmBooktitle{Proceedings of the 31st ACM International Conference on
	Multimedia (MM '23), October 29-November 3, 2023, Ottawa, ON, Canada}
\acmPrice{15.00}
\acmDOI{10.1145/3581783.3612005}
\acmISBN{979-8-4007-0108-5/23/10}
\begin{document}

\title{Multi-Spectral Image Stitching via Spatial Graph Reasoning}
%
\author{Zhiying Jiang}
\affiliation{%
	\institution{Dalian University of Technology}
	\city{}
	\country{}}
\email{zyjiang0630@gmail.com}

\author{Zengxi Zhang}
\affiliation{%
	\institution{Dalian University of Technology}
	\city{}
	\country{}}
\email{cyouzoukyuu@gmail.com}

\author{Jinyuan Liu}
\affiliation{%
	\institution{Dalian University of Technology}
	\city{}
	\country{}}
\email{atlantis918@hotmail.com}

\author{Xin Fan}
\affiliation{%
	\institution{Dalian University of Technology}
	\city{}
	\country{}}
\email{xin.fan@dlut.edu.cn}

\author{Risheng Liu}
\authornote{Corresponding author: Risheng Liu.}
\affiliation{%
	\institution{Dalian University of Technology}
	\institution{Peng Cheng Laboratory}
	\city{}
	\country{}}
\email{rsliu@dlut.edu.cn}


\begin{abstract}
Multi-spectral image stitching leverages the complementarity between infrared and visible images to generate a robust and reliable wide field-of-view~(FOV) scene. The primary challenge of this task is to explore the relations between multi-spectral images for aligning and integrating multi-view scenes. Capitalizing on the strengths of Graph Convolutional Networks~(GCNs) in modeling feature relationships, we propose a spatial graph reasoning based multi-spectral image stitching method that effectively distills the deformation and integration of multi-spectral images across different viewpoints. To accomplish this, we embed multi-scale complementary features from the same view position into a set of nodes. The correspondence across different views is learned through powerful dense feature embeddings, where both inter- and intra-correlations are developed to exploit cross-view matching and enhance inner feature disparity. By introducing long-range coherence along spatial and channel dimensions, the complementarity of pixel relations and channel interdependencies aids in the reconstruction of aligned multi-view features, generating informative and reliable wide FOV scenes. Moreover, we release a challenging dataset named ChaMS, comprising both real-world and synthetic sets with significant parallax, providing a new option for comprehensive evaluation. Extensive experiments demonstrate that our method surpasses the state-of-the-arts.
	

\end{abstract}




\maketitle
\begin{figure}[!t]
	\centering
	\setlength{\tabcolsep}{1pt}
	\begin{tabular}{cccccccccccc}	
		\includegraphics[width=0.48\textwidth]{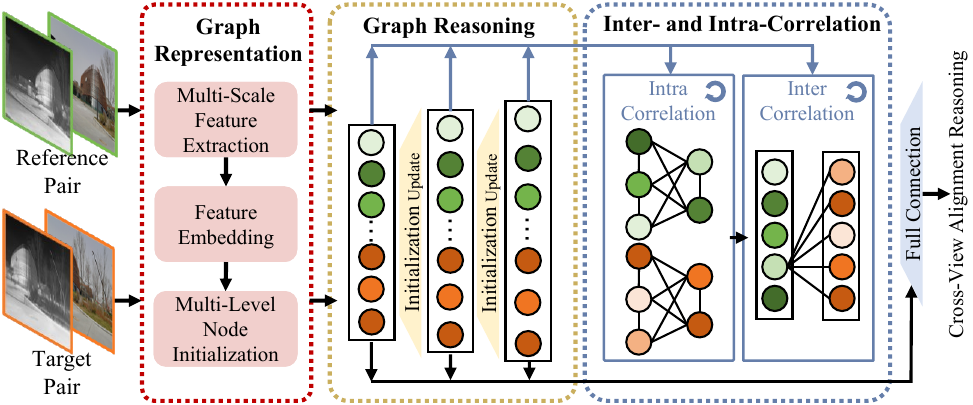}\\
	\end{tabular}\vspace{-1em}
	\caption{Overview of graph based alignment reasoning, in which inter- and intra-correlation are adopted to enhance the cross-view correspondence and the mono-view disparity. }\vspace{-2em}
	\label{fig:first_figure}
\end{figure}

\section{Introduction}
Image stitching refers to combining multiple overlapping images to reconstruct a wide field-of-view~(FOV) scene, which has been extensively employed in panoramic perception~\cite{gao2022review,Zhu_2022_CVPR}, vehicle navigation~\cite{koh2020real,chen2021semantic}, and virtual reality~\cite{chang2020virtual}. The majority of existing image stitching methods~\cite{gao2011constructing,lin2011smoothly,zaragoza2013projective} are tailored for visible images. However, in real-world scenarios, harsh imaging conditions, such as low light or adverse weather, introduce noise, occlusion, and information loss in the captured visible images, consequently undermining the effectiveness and robustness of the stitching algorithm.

In practice, the deployment of multi-spectral sensors enhances scene perception by capitalizing on their complementary imaging characteristics~\cite{liu2023holoco,liu2022unified,liu2022learn,liu2021smoa,liu2020bilevel}. 
Firstly, standard cameras are sensitive to visible light, the captured visible images accurately represent the colors and details of objects or scenes as they appear under the current lighting conditions~\cite{liu2019knowledge,jiang2022bilevel}. In contrast, thermal cameras detect invisible infrared radiation and convert it into visible images. Infrared images obtained from thermal cameras display varying temperature levels as distinct shades of gray, effectively capturing structural information while potentially lacking in textural details~\cite{Zhou_2022_CVPR,AdaptivePan2022MM}. As a result, as the most commonly used multi-spectral data, infrared and visible images based stitching holds a significant research value for achieving robust panoramic perception.

The intuitive process of multi-spectral image stitching involves fusing the multi-spectral images first, and then applying the conventional stitching algorithm to the latent fused results. However, the fusion of multi-spectral images essentially involves the recombination of appearances, which potentially diminishes the distinct characteristics presented by different modalities. Besides, the accumulation of deviations interferes with the alignment across different view positions, posing additional challenges to the conventional stitching performance. In other words, the cascaded fusion-stitching process is not ideally suited for addressing multi-spectral stitching.

The essence of multi-spectral image stitching lies in effectively harnessing the complementarity of infrared and visible images to facilitate the precise spatial alignment.
In this paper, we propose a spatial graph reasoning based method for multi-spectral image stitching. By leveraging local neighborhood information and graph topology, graph convolutional networks~(GCNs) prove remarkable efficacy in learning meaningful relations from the represented nodes and edges. In our case, the multi-scale complementary features of infrared and visible images in identical view position are sampled into sets of nodes, while information is aggregated from local neighborhoods and node features are updated hierarchically. During the process of correlation reasoning, both inter-correlation and intra-correlation are considered. Inter-correlation attains spatial matching by calculating the correspondence from different viewpoints, whereas intra-correlation strengthens the disparities between feature nodes within a specific viewpoint via performing internal comparisons. To reconstruct the seamless and credible wide FOV scene, spatial graph convolution and channel graph convolution are embedded to explore the global relationship between pixels and interdependency across channels. In this way, rich representations from different dimensions provide complementary guidance for the generation of accurate and plausible panoramic scenes. The main contributions of this paper are as follows:
\begin{itemize}
	\item We propose a spatial graph reasoning based multi-spectral image stitching method. To the best of our knowledge, this is the first time to introduce the graph convolutional network into the relation modeling for multi-spectral cross-view alignment and integration.
	\item We facilitate the alignment reasoning with the combination of inter-correlation and intra-correlation, which respectively investigate cross-view matching and intensify feature disparities within a particular viewpoint.
	\item The pixel relations and channel interdependencies from spatial and channel graph convolutions are employed in the reconstruction phase, advancing the generation of informative and seamless wide FOV scene.
	\item 
	We release a challenging multi-spectral image stitching dataset, ChaMS, including both real-world and synthetic sets. The stitching ground truth is provided for the synthetic pairs, offering a new option for evaluation. Extensive evaluation demonstrates the superiority of the proposed method.
\end{itemize}
 
\section{Related Work}
\subsection{Image Stitching}
Previous feature detection based methods~\cite{brown2007automatic,adel2014image} employed feature descriptors such as HOG~\cite{lowe2004distinctive}, SIFT~\cite{lowe2004distinctive}, and RANSAC~\cite{fischler1981random} to enable automatic stitching. However, due to the inherent limitations of descriptors in feature representation, these methods often encountered ghosting effects or other interferences. To alleviate adverse effects, Gao~\emph{et al.}~\cite{gao2011constructing} segmented the paired images into foreground and background, and utilized a distinct dual homography to match them. Lin~\emph{et al.}~\cite{lin2011smoothly} developed a smoothly varying affine~(SVA) transformation to address the variable parallax. Simultaneously, Zaragoza~\emph{et al.}~\cite{zaragoza2013projective} proposed the as-projective-as-possible~(APAP), allowing local nonprojective deviations while maintaining a global projective constraint. Based on a mesh framework, Zhang~\emph{et al.}~\cite{zhang2016multi} improved alignment and regularity constraints to support wide baselines and non-planar structures. Chen~\emph{et al.}~\cite{chen2016natural} implemented a grid mesh to direct the warping in local warp models and employed a global similarity prior~(GSP) for constraining the overall transformation to minimize local distortion.
In order to reduce dependence on feature detection, Lin~\emph{et al.}~\cite{lin2017direct} focused on minimizing pixel intensity differences rather than the Euclidean distance between corresponding features. Additionally, Lee~\emph{et al.}~\cite{lee2020warping} divided source images into superpixels and adaptively warped them using the optimal homography, further refining the stitching process.

Recent research has explored the deep learning to tackle the challenges of image stitching. DeTone~\emph{et al.}~\cite{detone2016deep} introduced the pioneering homography estimation network, while Nguyen~\emph{et al.}~\cite{nguyen2018unsupervised} devised an unsupervised learning approach for planar homography estimation. Nie~\emph{et al.}~\cite{nie2020view} proposed a content revision network to address the stitching seam and then developed an ablation constraint to reconstruct the wide FOV from feature to pixel~\cite{nie2021unsupervised}. Song~\emph{et al.}~\cite{song2022weakly} presented a weakly supervised learning method for fisheye panorama generation. Despite the effectiveness of these methods, environmental factors can adversely affect the captured visible images, subsequently hindering the generalizability of them.

\subsection{Graph Neural Networks}
Graph Convolutional Networks~(GCNs) have demonstrated significant potential in various tasks by modeling complex relationships to effectively exploit the underlying structure of images. Essentially, GCNs operate as message-exchanging systems, iteratively generating node representations by taking into account neighboring nodes via a differentiable combining process. Notable works in computer vision include Yan~\emph{et al.}~\cite{yan2018spatial} proposed spatial-temporal GCNs for skeleton-based action recognition, which leverages both spatial and temporal dependencies for efficient recognition of human actions. Wang~\emph{et al.}~\cite{gu2019scene} employed a relation-infused graph attention to generate scene graphs for image reconstruction. In 3D object detection, Chen~\emph{et al.}~\cite{chen2022graph} aggregated multi-view image information using graph structure learning. In the context of multi-spectral image stitching, challenges arise from effectively ensembling complementary information from infrared and visible images, as well as reasoning correspondence across different view positions. To tackle these issues, we develop a hierarchical mechanism to sequentially exploit multi-level feature relations and reason graph models from both inter and intra aspects, enhancing the spatial matching of multi-spectral images.
\begin{figure*}[t]
	\centering
	\setlength{\tabcolsep}{1pt}
	\begin{tabular}{cccccccccccc}	
		\includegraphics[width=1\textwidth]{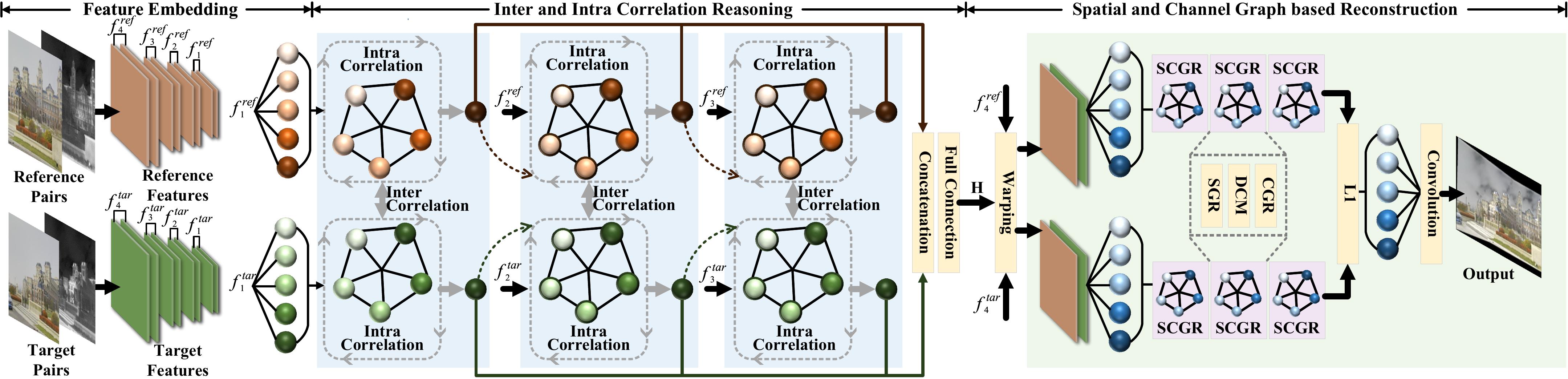}\\
	\end{tabular}\vspace{-1em}
	\caption{Workflow of the proposed method. It involves feature embedding, correlation reasoning, and reconstruction phases. The inter and intra correlation are employed for more accurate spatial relation reasoning. The long-range coherence in spatial and channel dimensions are investigated for the comprehensive and plausible wide FOV reconstruction.  }
	\label{fig:workflow}
\end{figure*}

\begin{figure}[t]
	\centering
	\setlength{\tabcolsep}{1pt}
	\begin{tabular}{cccccccccccc}	
		\includegraphics[width=0.46\textwidth]{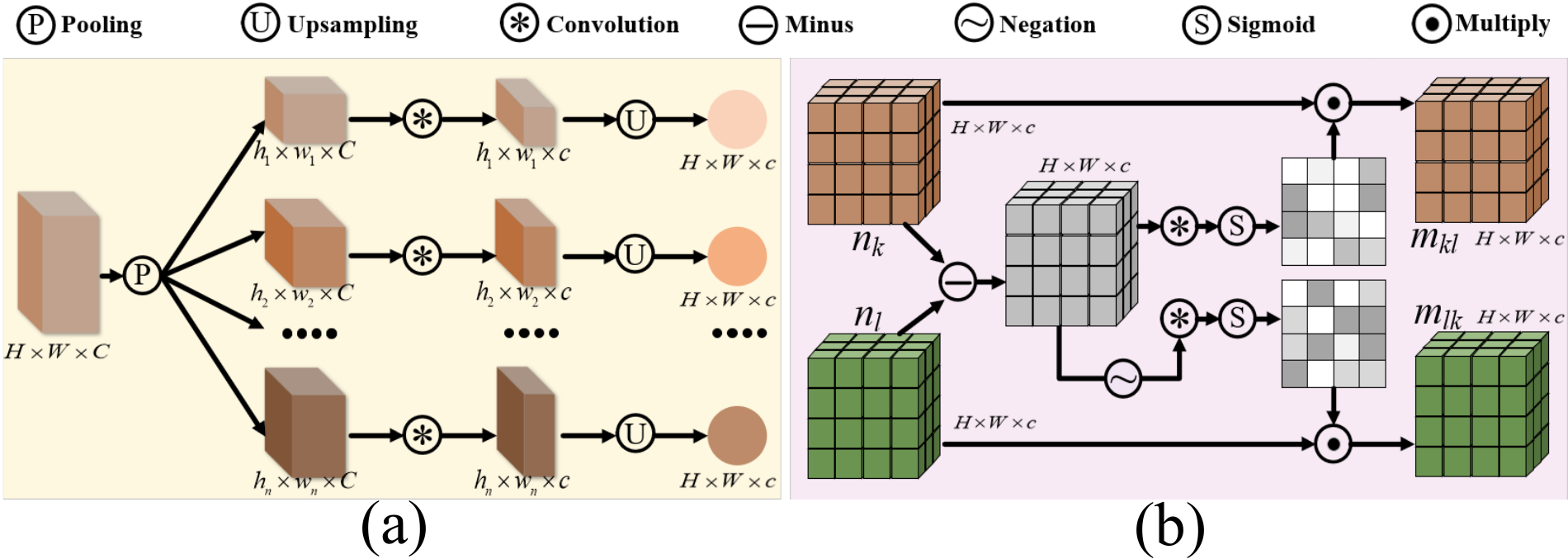}\\
	\end{tabular}\vspace{-1em}
	\caption{(a) shows the detailed process of feature embedding, while~(b) demonstrates the comprehensive message passing between any pair of nodes.}
	\label{fig:edge_embedding}\vspace{-2em}
\end{figure}

\subsection{Infrared and Visible Image Datasets}
The progress in multi-spectral sensing has given rise to an array of infrared and visible image datasets. The initial FLIR~\footnote{https://www.flir.com/oem/adas/adas-dataset-form/} dataset exhibited misalignment between paired infrared and visible images, rendering it unsuitable for stitching assessment. The TNO~\cite{toet2017tno} dataset, captured in various military and surveillance settings, contains ambiguous appearances and noise interference. RoadScene~\cite{xu2020u2fusion} dataset enhanced the alignment accuracy of FLIR, releasing~$221$ infrared-visible image pairs focused primarily on vehicles and pedestrians as objects and targets.~$\text{M}^3\text{FD}$~\cite{Liu_2022_CVPR} dataset presented a comprehensive infrared-visible benchmark for multi-modality image detection, covering a majority of road driving scenarios and featuring numerous instances of dynamic blurring. More recently, MSIS~\cite{jiang2022towards} dataset introduced a multi-spectral image stitching-applicable dataset, consisting of~$381$ infrared-visible image pairs with diverse overlapping rates and complex content. However, as the source data are captured in real-world settings, the ground truth for the corresponding wide FOV scenes is unavailable, limiting the quantitative evaluation.


\section{The Proposed Method}
Given two pairs of multi-spectral images~$\{x_{ir}^{ref},x_{vis}^{ref}\}, \{x_{ir}^{tar},x_{vis}^{tar}\}$, $ir$ and~$vis$ denote the infrared and visible images respectively, while $tar$ and~$ref$ means the target pairs and reference counterpart. Multi-spectral image stitching benefits from the integration of infrared and visible information, generating a more accurate and reliable wide field-of-view~(FOV) scene. Correlation based cross-view alignment and seamless panorama reconstruction are two crucial steps of this task. Leveraging the power of graph convolutional networks to characterize global-local spatial information, we propose to reason the spatial correlation using a graph structure. Initially, the complementary features from a specific viewpoint are embedded into a set of node vectors. Plane transformations across different viewpoints are estimated using a progressive mechanism. In each stage, both intra- and inter-correlations are investigated, with prior reasoning enabling a pre-update of subsequent reasoning. The resulting three-scale reasoning is concatenated to regress the desired homography. During the panorama reconstruction phase, Spatial and Channel Graph Reasoning~(SCGR) are employed to explore pixel relations and channel interdependencies. Given the distinct characteristics of multi-spectral images, an L1-norm based integration is utilized, ensuring the preservation of salient structural information and significant details. After fusing the complementary representations, a comprehensive and plausible wide FOV can be reconstructed. An overview of our method is shown in Fig.~\ref{fig:workflow}. Then we provide a detailed introduction of each module.

\subsection{Feature Embedding}
We first employ two modality-specific VGG-16~\cite{simonyan2014very,liu2022twin,jiang2022target} to independently encode the features of multi-spectral images at different viewpoints. Following this, the features of the corresponding layers are concatenated, resulting in two feature pyramids with four-scale features for the reference and target viewpoints. To fully distill complementary information from the multi-spectral images, we are supposed to learn powerful representations so as to realize more accurate alignment with rich, complementary information from multi-level features. We leverage the pooling operation followed by a convolutional layer and an upsample layer to extract multi-scale features of different viewpoints as the initial node representations, expressed as:
\begin{equation}
	n^{(i)}=\mathcal{U}(Conv(\mathcal{P}(f^{(i)}))),\label{feature_embed}
\end{equation}
where~$f^{(i)}$ denote the features need to be distilled,~$i$ is the feature level,~$\mathcal{P}(\cdot)$ is the pooling operation and~$\mathcal{U}(\cdot)$ means the upsample layer which ensures the embedded multi-scale feature maps hold the same size. The detailed illustration is shown in Fig.~\ref{fig:edge_embedding} (a).
\subsection{Inter and Intra Correlation Reasoning}
In order to explicitly reason on high-level relations over infrared and visible images for better alignment, each node aggregates feature messages from all its neighboring nodes. Based on the initially embedding nodes obtained above, we utilize the parametric edges between the nodes to represent the spatial correlation, and adopt this parameter to update the initial nodes. As shown in Fig.~\ref{fig:edge_embedding}~(b), the message between any pairs of nodes is bidirectional. For the messages~$m_{kl}, m_{lk}$ passed from all neighboring nodes~$n_k$to~$n_{l}$, it can be determined as:
\begin{equation}
	\begin{split}
	m_{kl}=n_k*[Sig(Conv(n_k-n_l))],\\
	m_{lk}=n_l*[Sig(Conv(n_l-n_k))],\label{mlk}
\end{split}
\end{equation}
where~$Sig(\cdot)$ is the sigmoid function,~$*$ denote the element-wise multiply. And the update of the related node can be expressed as:
\begin{equation}
	n_k=n_k+m_{lk},\ \ n_l=n_l+m_{kl}.
\end{equation}
The relation of multi-view complementary feature are reasoned from two aspects, i.e., inter and intra correlations. The inter correlation focus on cross-view relation, where~$n_k$ and~$n_l$ in Eq.~\eqref{mlk} belong to different sets of node. Conversely, as multi-spectral complementary features are embedded within the same set, intra correlations are used to enhance the inner feature disparity among nodes in the same set. For the whole reasoning of a certain level features, intra correlation is calculated twice, followed by two rounds of inter correlation computation.

Based on the four-scale feature pyramid, correlation reasoning is conducted with a progressive mechanism. We start from the smallest scale features~$\{f^{ref}_1,\ f^{tar}_1\}$, and the updated nodes within inter and intra correlation in this scale can be mapped into two view-specific guidance nodes~$n_{g1}^{ref},n_{g1}^{tar}$ to incorporate the subsequent node initialization. Concretely, the guidance nodes~$n_{g1}^{ref},n_{g1}^{tar}$ can be formulated as:
\begin{equation}
	n_{g1}=Conv(Concat(n_1,\cdot\cdot\cdot, n_j)),\ j=N,
\end{equation}
where~$n_1,\cdot\cdot\cdot, n_j$ are nodes after two round of intra and inter update,~$N$ denotes the number of nodes. The initialization nodes~$\breve{n}^{(2)}$ of~$\{f^{ref}_2,\ f^{tar}_2\}$ are defined as:
\begin{equation}
	\breve{n}^{(2)}=Sig(\mathcal{A}(n_{g1}))*n^{(2)}+n^{(2)},
\end{equation}
where~$n^{(2)}$ can be obtained in Eq.~\eqref{feature_embed},~$\mathcal{A}(\cdot)$ denotes the global average pooling. In this way, the feature embedding of~$\{f^{ref}_2,\ f^{tar}_2\}$ fuses the preceding graph information, and the relation reasoning in this scale is same as the preceding level.

Through the progressive mechanism, three sets of guidance nodes can be obtained, i.e.,~$\{n_{g1}^{ref}, n_{g1}^{tar}\}, \{n_{g2}^{ref}, n_{g2}^{tar}\}, \{n_{g3}^{ref}, n_{g3}^{tar}\}$. The homography for multi-spectral, multi-view alignment can be derived from these nodes, which can be formulated as:
\begin{equation}
	H=\mathcal{R}(Concat(n_{g1}^{ref}, n_{g1}^{tar}, n_{g2}^{ref}, n_{g2}^{tar}, n_{g3}^{ref}, n_{g3}^{tar})),
\end{equation}
where~$\mathcal{R}(\cdot)$ denotes full connection. The different view positions can be transformed into the same plane by warping the target feature~$f^{tar}_4$ into the reference~$f^{ref}_4$.
\begin{figure}[t]
	\centering
	\setlength{\tabcolsep}{1pt}
	\begin{tabular}{cccccccccccc}	
		\includegraphics[width=0.46\textwidth]{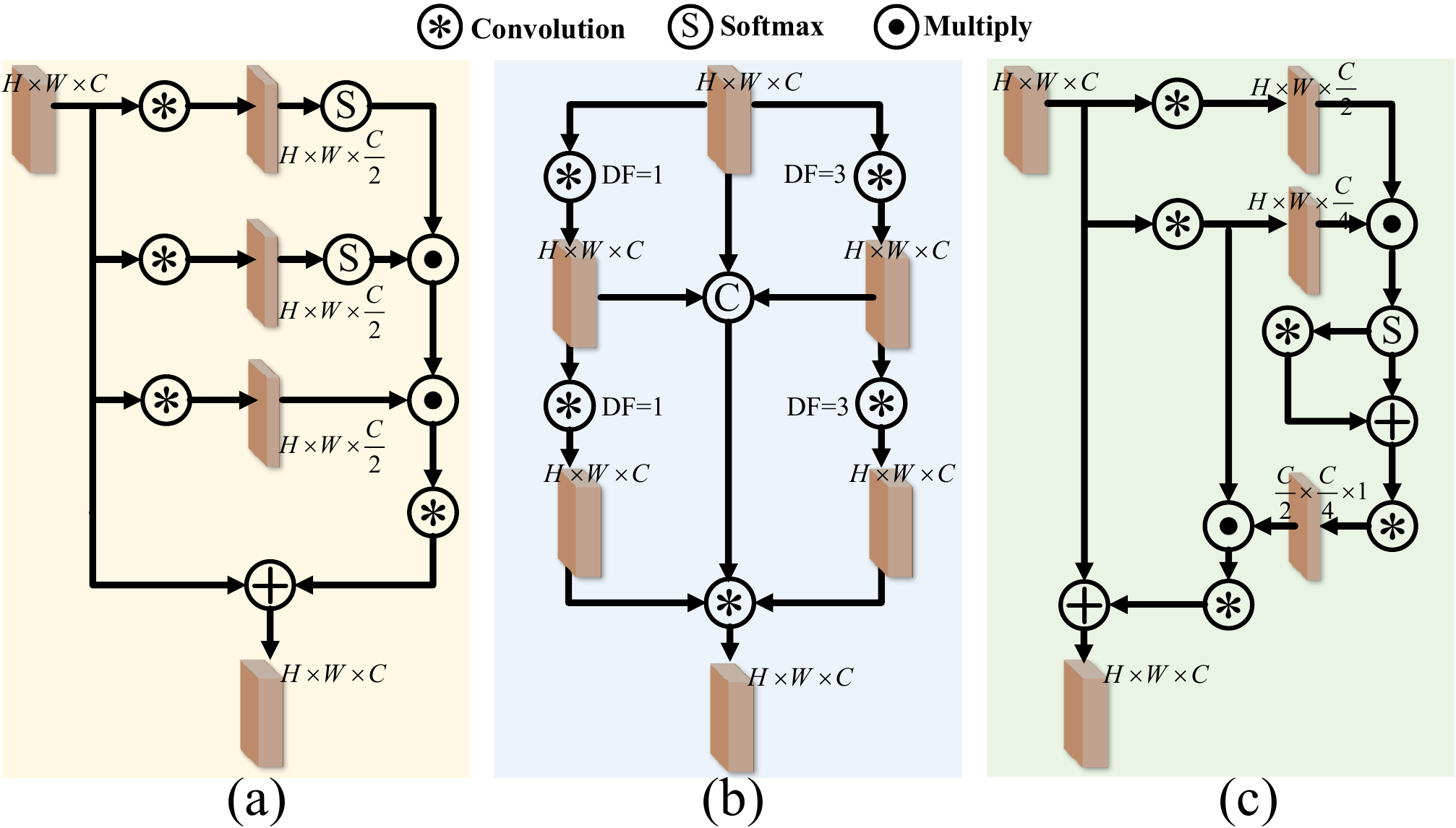}\\
	\end{tabular}\vspace{-1em}
	\caption{ Detailed architecture of the sub-modules used in the reconstruction phase. (a) spatial graph reasoning~(SGR), (b) dilated convolutional module~(DCM) and (c) channel graph reasoning~(CGR). }
	\label{fig:spatial_channel_graph}\vspace{-1em}
\end{figure}
\subsection{Spatial and Channel Graph based Reconstruction}
During the reconstruction of wide FOV scenes, challenges such as ghosting, overlapping, and distortion often arise, since the vanilla CNNs only model local spatial information, ignoring the long-range dependencies of texture. To address these issues and generate an accurate and reliable panorama, we propose leveraging spatial long-range coherence to improve the perception of the warped multi-viewpoint features, thereby enhancing the accuracy of multi-viewpoint content fusion. Moreover, given the complementarity of multi-spectral imaging, we integrate information from both infrared and visible images during the stitching process, ensuring the preservation of distinct information of multi-spectral images. 

Specifically, we develop a Spatial and Channel Graph based Reconstruction~(SCGR), where information from multiple dimensions is exploited, including pixel-wise global spatial relations and channel-wise global interdependencies. In practice, the SCGR module is composed of Spatial Graph Reasoning~(SGR), Channel Graph Reasoning~(CGR), and a Dilated Convolutional Module~(DCM) to bridge SGR and CGR together. 
\subsubsection{Spatial Graph Reasoning} It explores the relation between one pixel and all pixels in the feature map. As shown in Fig.~\ref{fig:spatial_channel_graph}~(a), we first employ three~$1\times1$ convolution operations to reduce the channel number and use softmax to avoid numerical instabilities. The features after spatial reasoning is obtained with a~$1\times1$ convolution based weighting process, expressed as:
\begin{equation}
	f_{SGR}=f+Conv(\phi(Conv(f))*\phi(Conv(f))*Conv(f)),
\end{equation}
where~$\phi(\cdot)$ presents softmax operation,~$*$ is multiply operation. SGR produces coherent prediction covering all pixels, facilitating the extraction of long-range relations. 
\subsubsection{Dilated Convolutional Module} After the SGR, a dual-path dilated convolutional process is employed. One pathway comprises two conventional convolutional layers designed to capture small-scale spatial patterns, while the other consists of two dilated convolutional layers, which rapidly expand the receptive field. As illustrated in Fig.~\ref{fig:spatial_channel_graph}~(b), the output of this module is derived through the fusion of five distinct features using a~$1\times1$ convolutional operation, thereby effectively extracting multi-scale local spatial features.

\subsubsection{Channel Graph Reasoning} It investigates the interrelations among features within the channel dimension. As shown in Fig.~\ref{fig:spatial_channel_graph}~(c), we initially create latent channel correlation features~$f_c$ using two~$1\times1$ convolutions and softmax to realize the channel-wise information aggregation, expressed as:
\begin{equation}
	f_c=\phi(Conv(f)*Conv(f)).
\end{equation}
To reason the channel correlation, a 1D convolution is performed on the latent features, followed by a combination of hidden-to-output operations, as expressed:
\begin{equation}
	\begin{split}
		\breve{f}_c&=f_c+ Conv(f_c),\\
		f_{CGR}&=f+Conv(Conv(\breve{f}_c)*Conv(f)).
	\end{split}
\end{equation}
By incorporating CGR into the reconstruction module, we can effectively identify and understand the interdependencies between channels within the feature map.

In the proposed method, dual cascaded paths, each with three SCRM components, are designed to effectively capture the relationships and interdependencies between both infrared and visible images. To preserve the distinct information of multi-spectral images, we integrate the complementary reasoning via L1-norm~\cite{li2018densefuse} and reconvert the embedded features into the desired image with wide FOV and informative appearance.
\vspace{-.5em}
\subsection{Loss Function}
Considering the proposed method, two learnable subnetworks need to be trained: the correlation reasoning network and the spatial and channel graph based reconstruction network. For the correlation reasoning network, the loss function is employed to constrain the alignment performance, taking both infrared and visible images into account. A comparison is conducted between the common regions of the warped target features and their corresponding reference counterparts, expressed as:
\begin{equation}
	\begin{split}
		\mathcal{L}_H=&|\mathbf{S}(x^{ref}_{vis})-\mathcal{S}(\mathcal{W}(x^{tar}_{vis},H))|_1+|\mathbf{S}(x^{ref}_{ir})-\mathcal{S}(\mathcal{W}(x^{tar}_{ir},H))|_1,
	\end{split}
\end{equation}
where~$H$ is the regressed homography based on the reasoned correlation,~$\mathcal{W}(\cdot)$ represents the warping operation used to transform the target view onto the reference plane,~$\mathcal{S}(\cdot)$ denotes the extraction of common regions. For the reconstruction network, we considered three types of loss functions: L1 constraints on the seam regions, structural similarity constraints on the content, and perceptual constraints. Initially, we acquired masks for the seam regions~\cite{jiang2022towards} corresponding to the two view positions and employed the L1-norm to calculate the loss, expressed as:
\begin{equation}
	\begin{split}
		\mathcal{L}_{1}&=\lambda_1|x_{out}\odot S_1-\breve{x}_{ir}^{ref}\odot S_1|_1+\lambda_2|x_{out}\odot S_2-\breve{x}_{ir}^{tar}\odot S_2|_1\\
		&+\lambda_2|x_{out}\odot S_1-\breve{x}_{vis}^{ref}\odot S_1|_1+\lambda_2|x_{out}\odot S_2-\breve{x}_{vis}^{tar}\odot S_2|_1,
	\end{split}
\end{equation}
where~$x_{out}$ is the reconstructed result,~$S_1, S_2$ denote the seam masks in reference and target viewpoints.~$\odot$ presents the multiply operation,~$\breve{x}$ means the warped corresponding image.~$\lambda_1=1, \lambda_2=1.5$ are balancing parameters. The structural similarity constraints are enforced by identifying the common regions between the stitched result and the scenes from distinct view positions, and applying the SSIM~(Structural Similarity Index Measure) to the shared regions, which can be formulated as:
\begin{equation}
	\begin{split}
		\mathcal{L}_{ssim}=\lambda_3 (&1-\text{SSIM}(x_{out}\odot\mathbf{C}_1,\breve{x}_{ir}^{ref}\odot \mathbf{C}_1)+\\
		&1-\text{SSIM}(x_{out}\odot\mathbf{C}_2,\breve{x}_{ir}^{tar}\odot \mathbf{C}_2))\\
		+\lambda_4 (&1-\text{SSIM}(x_{out}\odot\mathbf{C}_1,\breve{x}_{vis}^{ref}\odot \mathbf{C}_1)+\\
		&1-\text{SSIM}(x_{out}\odot\mathbf{C}_2,\breve{x}_{vis}^{tar}\odot \mathbf{C}_2)),
	\end{split}
\end{equation}
where~$C_1, C_2$ denote the content masks in reference and target viewpoints.~$\lambda_3=10, \lambda_4=15$ are balancing parameters. Besides, the perceptual loss constrains the consistency of features within the shared content.:
 \begin{equation}
 	\begin{split}
 		\mathcal{L}_{per}=\lambda_5 &(|\mathcal{V}(x_{out}\odot\mathbf{C}_1)-\mathcal{V}(\breve{x}_{ir}^{ref}\odot\mathbf{C}_1)|_2+\\
 		&|\mathcal{V}(x_{out}\odot\mathbf{C}_2)-\mathcal{V}(\breve{x}_{ir}^{tar}\odot \mathbf{C}_2)|_2)\\
 		+\lambda_6 &(|\mathcal{V}(x_{out}\odot\mathbf{C}_1)-\mathcal{V}(\breve{x}_{vis}^{ref}\odot \mathbf{C}_1)|_2+\\
 		&|\mathcal{V}(x_{out}\odot\mathbf{C}_2)-\mathcal{V}(\breve{x}_{vis}^{tar}\odot \mathbf{C}_2)|_2),
 	\end{split}
 \end{equation}
$\mathcal{V}(\cdot)$ is the VGG feature extractor.~$\lambda_5=1e^{-3}, \lambda_6=1e^{-3}$. Accordingly, the total loss of the proposed method is as follows:
\begin{equation}
	\mathcal{L}_{total}=\mathcal{L}_H+\mathcal{L}_1+\mathcal{L}_{ssim}+\mathcal{L}_{per}.
\end{equation}

\section{The Proposed Dataset}
To comprehensively assess multi-spectral image stitching, we release a new challenging dataset, ChaMS, comprised of both a real-world set~(ChaMS-Real) and a synthetic set~(ChaMS-Syn). The capture of real-world data is facilitated with a multi-spectral sensor module equipped with infrared and visible cameras. We focus on large baseline scenes and multiple angle variations to simulate challenging scenarios with significant parallax, where the depth difference undermines transformations across multiple viewpoints. To address the registration between infrared and visible images, which arises due to the different resolution and relative position, we employ manual registration based on the intrinsic and extrinsic parameters of the cameras. In this way, we obtained a total of~$593$ pairs of infrared-visible images applicable for stitching.

Acquiring ground truth for multi-spectral stitching in the real world is extremely difficult. Therefore, based on the RoadScene dataset~\cite{xu2020u2fusion}, we synthesized a ground truth available supervised dataset. First, we extract~$224\times224$ image patches at random from the original RoadScene dataset, which serve as reference images. To create stitchable pairs under different baseline conditions, we randomly shift the four vertices of the reference image and reshape the polygon formed by the modified vertices into a~$224\times224$ rectangular image patch, which serves as the target images. The region formed by the vertices before and after shifting represents the ground truth for image stitching. Ultimately, we synthesize a total of~$10,500$ pairs of multi-spectral images accompanied by ground truth for training and evaluation purposes. Table.~\ref{tab:dataset_comparison} presents a comparison between the proposed ChaMS and the MSIS dataset.
\vspace{-0.5em}	
\begin{table}[h]
	\begin{center}
		\footnotesize
		\setlength\tabcolsep{3pt}
		\centering
		\caption{ Comparison between the proposed ChaMS~(including ChaMS-Real and ChaMS-Syn) with the MSIS~\cite{jiang2022towards}.}
		\label{tab:dataset_comparison}\vspace{-1em}	
		\begin{tabular}{p{1.3cm}<{\centering}|p{1cm}<{\centering}|p{1cm}<{\centering}|p{1cm}<{\centering}|p{1cm}<{\centering}|p{1cm}<{\centering}}\hline			Dataset&Img pairs&Resolution&Color&Parallax&Annotation\\\hline
			MSIS&$381$&$896\times640$&\ding{51}&Horizontal&\ding{55}\\
			ChaMS-Real&$593$&$1152\times832$&\ding{51}&Arbitrary&\ding{55}\\
			ChaMS-Syn&$10,500$&$224\times224$&\ding{51}&Arbitrary&\ding{51}\\\hline	
		\end{tabular}\vspace{-1.5em}		
	\end{center}
\end{table}
\begin{figure*}[]
	\centering
	\setlength{\tabcolsep}{1pt}
	\begin{tabular}{cccccccccccc}	
		\includegraphics[width=0.05\textwidth,height=0.06\textheight]{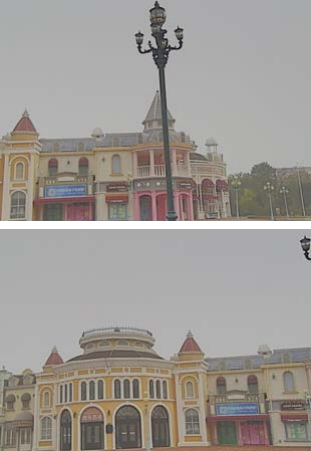}
		&\includegraphics[width=0.05\textwidth,height=0.06\textheight]{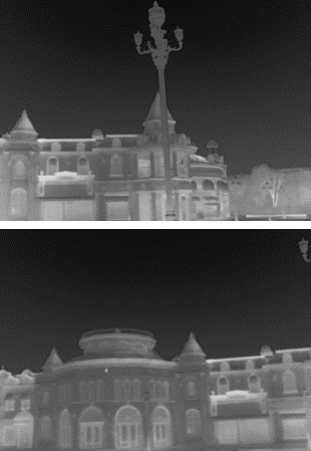}
		&\includegraphics[width=0.14\textwidth,height=0.06\textheight]{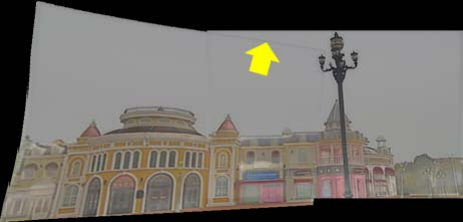}
		&\includegraphics[width=0.14\textwidth,height=0.06\textheight]{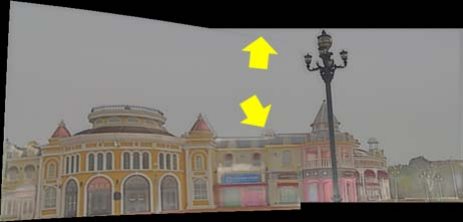}
		&\includegraphics[width=0.14\textwidth,height=0.06\textheight]{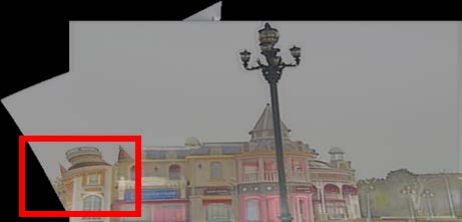}
		&\includegraphics[width=0.14\textwidth,height=0.06\textheight]{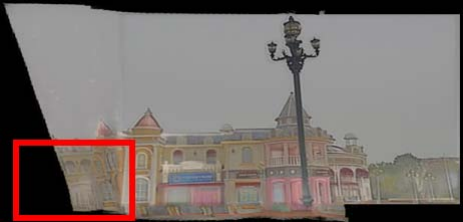}
		&\includegraphics[width=0.14\textwidth,height=0.06\textheight]{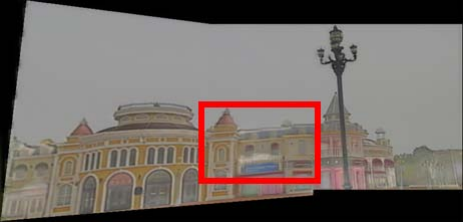}
		&\includegraphics[width=0.14\textwidth,height=0.06\textheight]{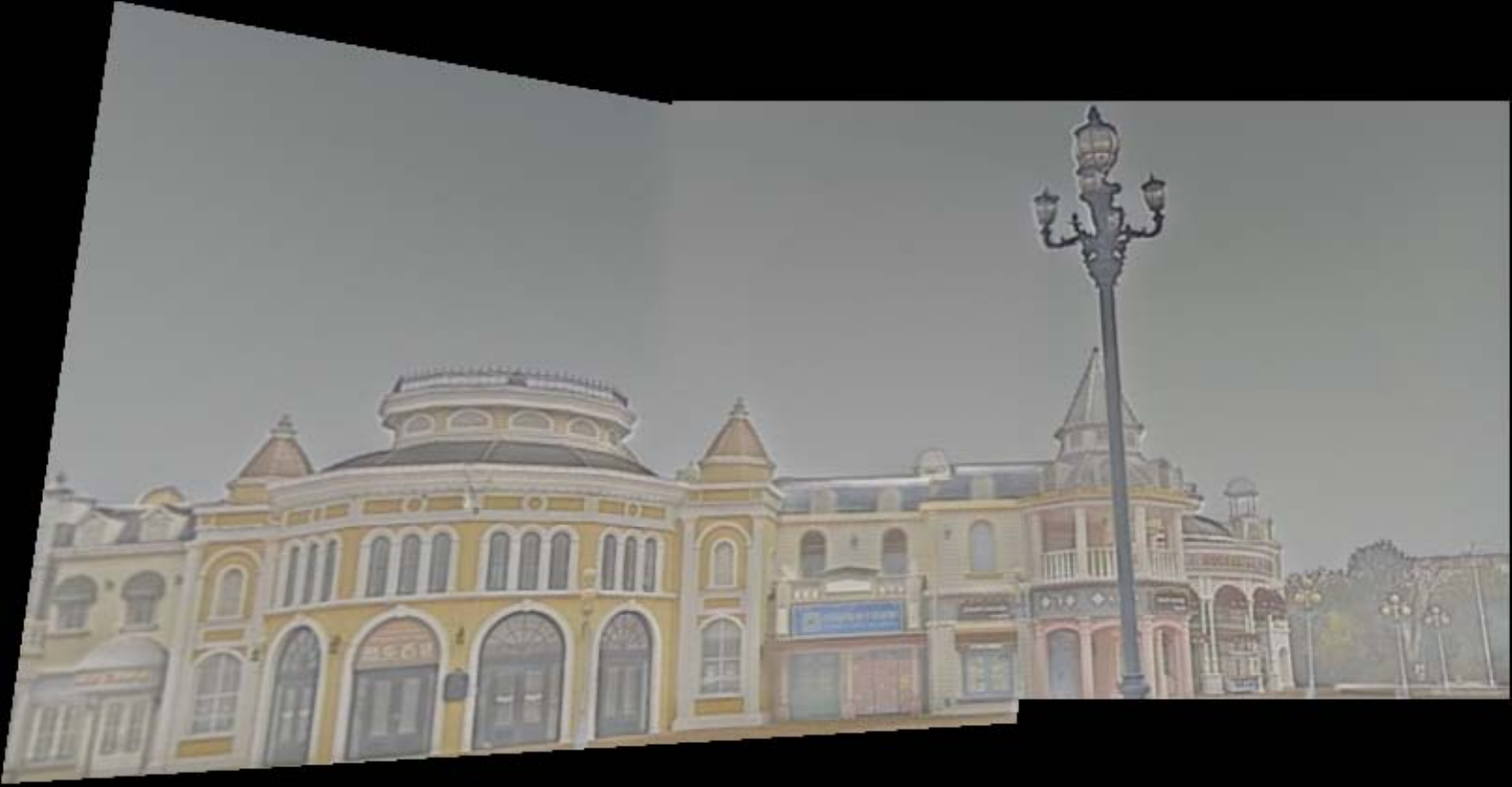}\\
		\includegraphics[width=0.05\textwidth,height=0.06\textheight]{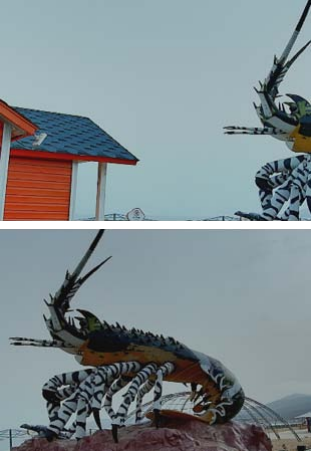}
		&\includegraphics[width=0.05\textwidth,height=0.06\textheight]{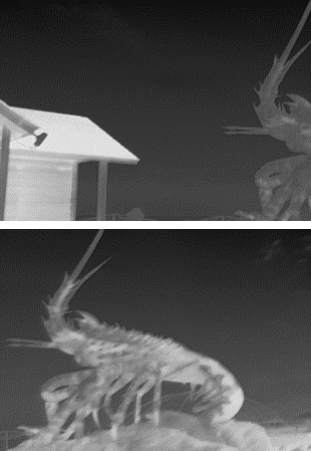}
		&\includegraphics[width=0.14\textwidth,height=0.06\textheight]{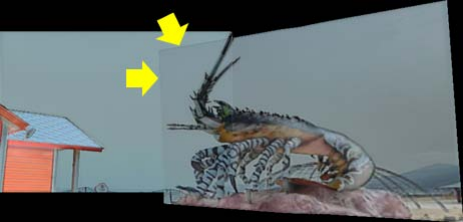}
		&\includegraphics[width=0.14\textwidth,height=0.06\textheight]{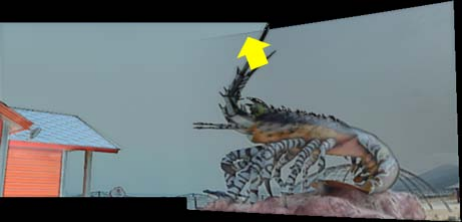}
		&\includegraphics[width=0.14\textwidth,height=0.06\textheight]{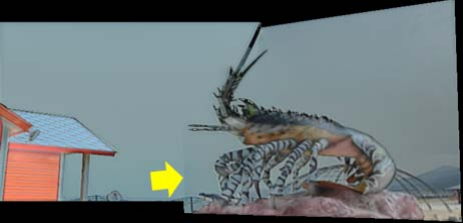}
		&\includegraphics[width=0.14\textwidth,height=0.06\textheight]{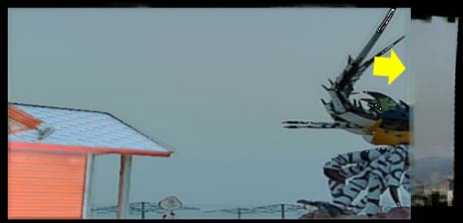}
		&\includegraphics[width=0.14\textwidth,height=0.06\textheight]{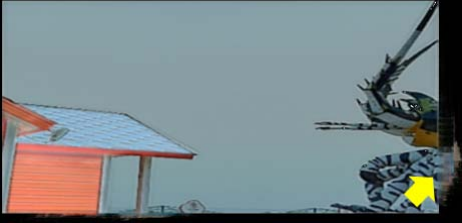}
		&\includegraphics[width=0.14\textwidth,height=0.06\textheight]{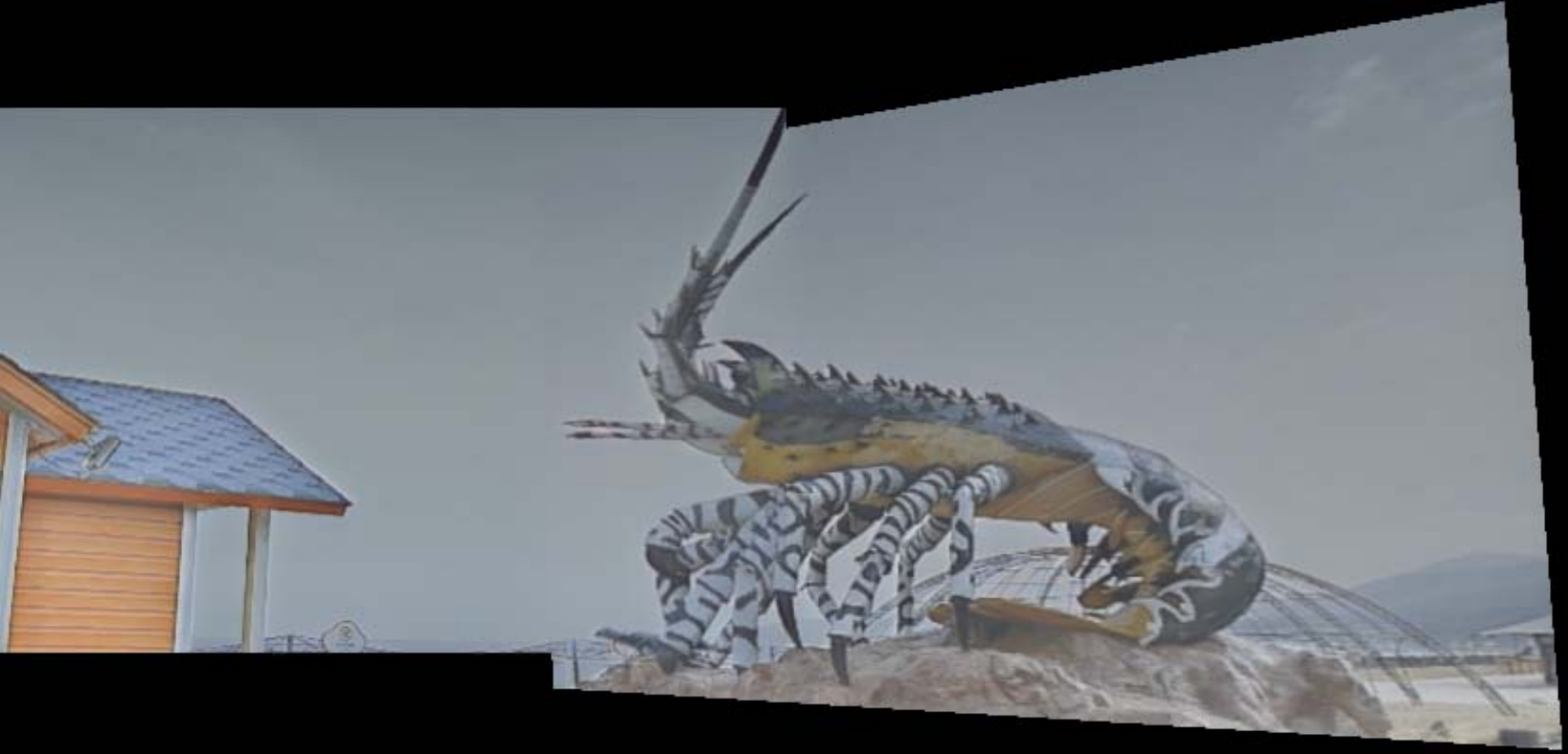}\\
		VIS&IR&APAP&SPW&WPIS&VFIS&RSFI&Ours\\
	\end{tabular}\vspace{-1.1em}
	\caption{The first row exhibits the visual comparison on the MSIS dataset, and the second row is on ChaMS-Real dataset. VIS and IR denote the input visible and infrared images.}
	\label{fig:vis_MSIS_challenge}
\end{figure*}
\begin{figure*}[]
	\centering
	\setlength{\tabcolsep}{1pt}
	\begin{tabular}{cccccccccccc}	
		\includegraphics[width=0.05\textwidth,height=0.080\textheight]{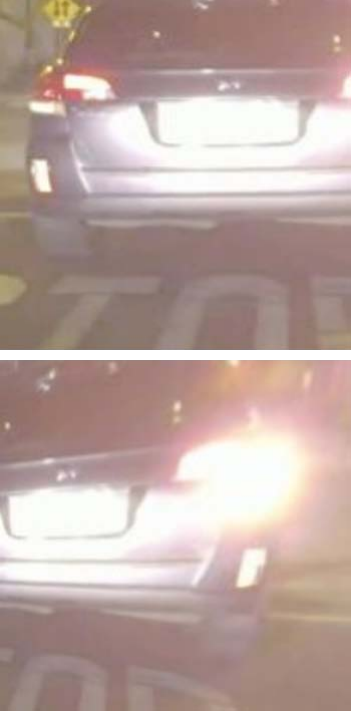}
		&\includegraphics[width=0.05\textwidth,height=0.080\textheight]{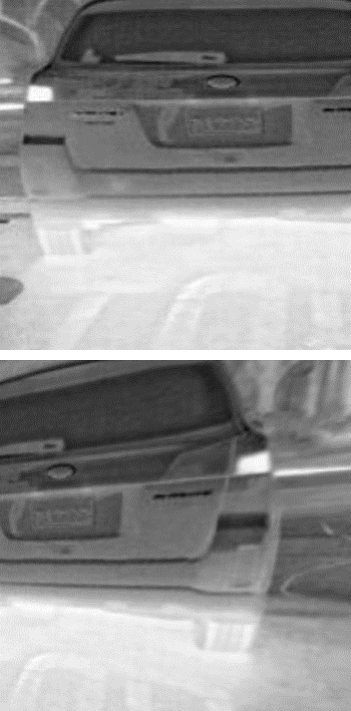}
		&\includegraphics[width=0.14\textwidth,height=0.080\textheight]{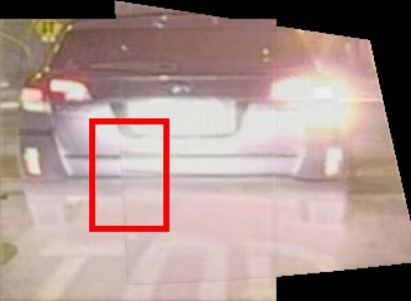}
		&\includegraphics[width=0.14\textwidth,height=0.080\textheight]{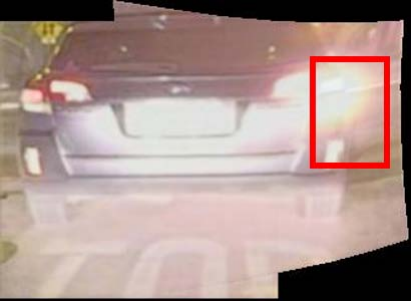}
		&\includegraphics[width=0.14\textwidth,height=0.080\textheight]{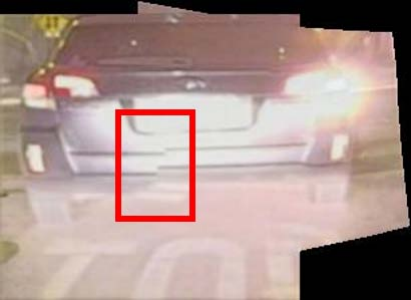}
		&\includegraphics[width=0.14\textwidth,height=0.080\textheight]{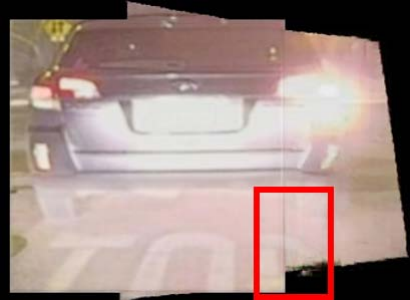}
		&\includegraphics[width=0.14\textwidth,height=0.080\textheight]{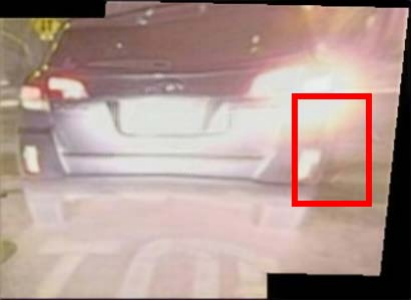}
		&\includegraphics[width=0.14\textwidth,height=0.080\textheight]{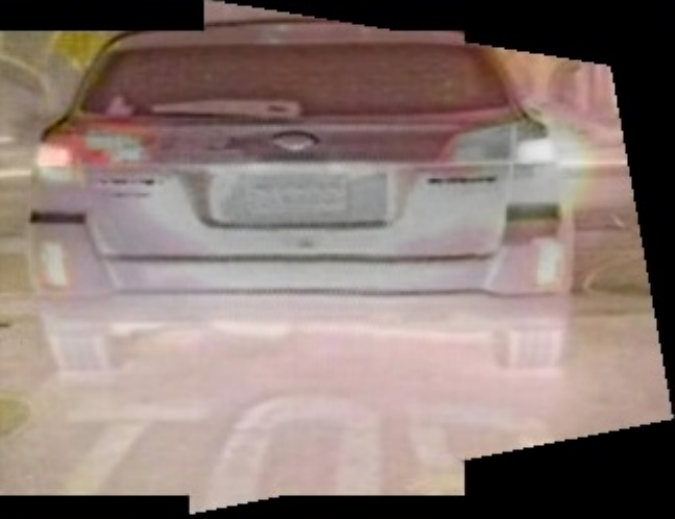}\\
		\includegraphics[width=0.05\textwidth,height=0.080\textheight]{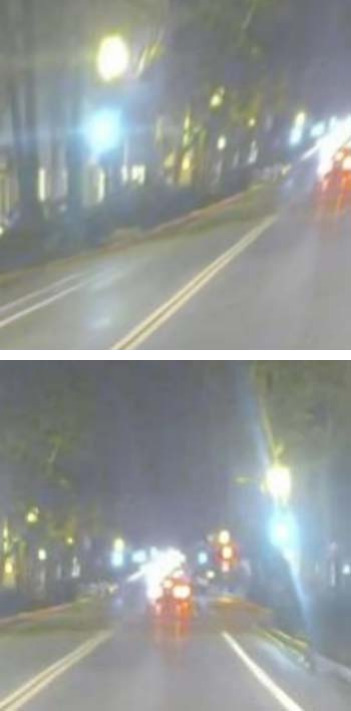}
		&\includegraphics[width=0.05\textwidth,height=0.080\textheight]{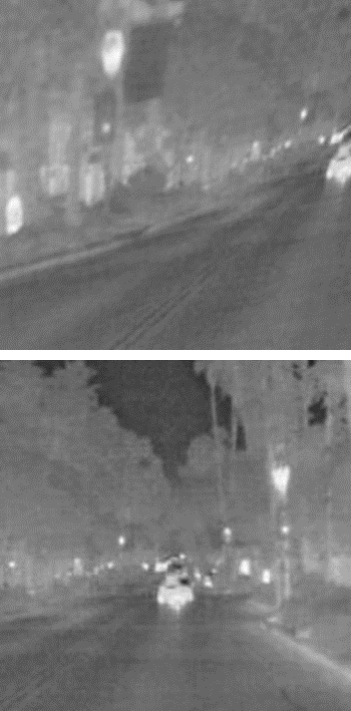}
		&\includegraphics[width=0.14\textwidth,height=0.080\textheight]{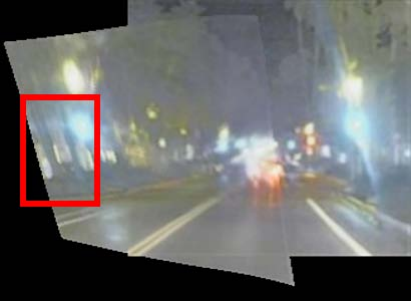}
		&\includegraphics[width=0.14\textwidth,height=0.080\textheight]{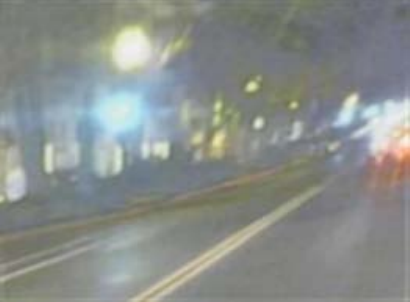}
		&\includegraphics[width=0.14\textwidth,height=0.080\textheight]{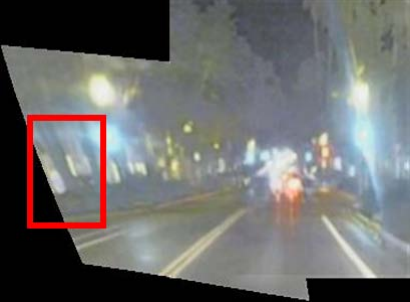}
		&\includegraphics[width=0.14\textwidth,height=0.080\textheight]{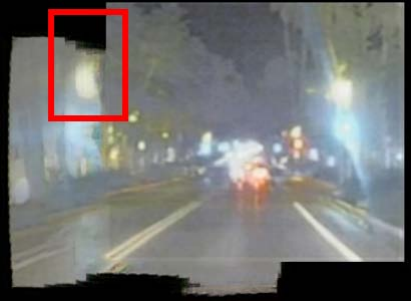}
		&\includegraphics[width=0.14\textwidth,height=0.080\textheight]{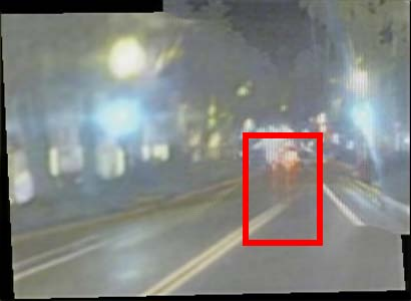}
		&\includegraphics[width=0.14\textwidth,height=0.080\textheight]{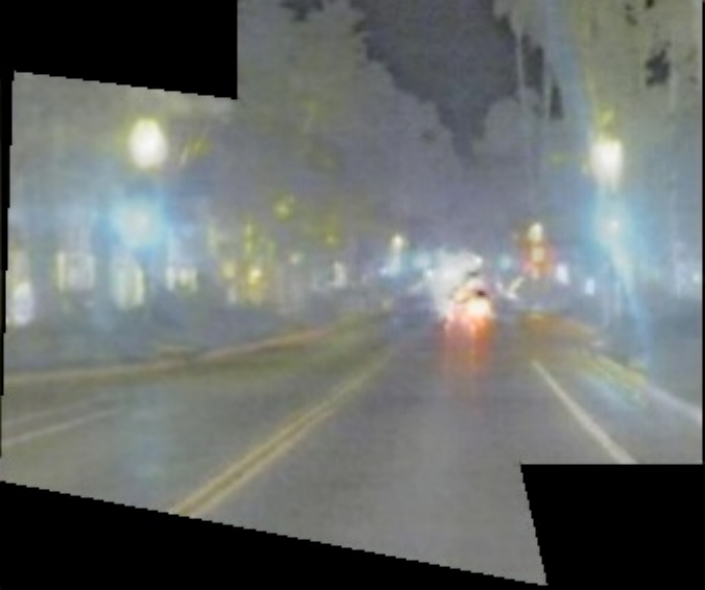}\\
		VIS&IR&APAP&SPW&WPIS&VFIS&RSFI&Ours\\
	\end{tabular}	\vspace{-1.1em}
	\caption{Visual comparison on the synthetic ChaMS-Syn dataset. 
}
		\vspace{-1em}
	\label{fig:vis_syn}
\end{figure*}

\section{Experimental Results}
\subsection{Implement Details}
The proposed method was implemented on Pytorch with an NVIDIA Geforce 3090 GPU. In the training process, we adopted~$10,000$ pairs of synthetic multi-spectral images from the proposed ChaMS-Syn as training data. In the initial stage, the correlation reasoning module was trained for~$150$ epochs using the Adam optimizer, featuring a decay rate of~$0.96$ and a learning rate of~$1e^{-4}$. Subsequently, the parameters of the correlation reasoning module were fixed, and the reconstruction module was trained independently for~$10$ epochs, adhering to the same hyperparameter configuration. Lastly, both the correlation reasoning and reconstruction modules were trained concurrently for~$50$ epochs, commencing with a learning rate of~$5e^{-5}$.

\vspace{-1em}
\subsection{Comparison with Existing Methods}
The proposed method generates wide FOV images that incorporate both infrared and visible information, thereby achieving the integration of multi-spectral and multi-view scenes during the reconstruction process. For a fair comparison, we first implemented several state-of-the-art infrared-visible image fusion algorithms, including FusionGAN~\cite{ma2019fusiongan}, GANMcC~\cite{ma2020ganmcc}, MFEIF~\cite{Liu2022learning}, RFN~\cite{li2021rfn}, and ReCoNet~\cite{huang2022reconet} to obtain the complementary and comprehensive images. Followed by the application of prolific stitching algorithms for stitching performance evaluation, including conventional based APAP~\cite{zaragoza2013projective}, SPW~\cite{liao2019single}, and WPIS~\cite{jia2021leveraging}, as well as two deep learning based methods, specifically VFIS~\cite{nie2020view} and RSFI~\cite{nie2021unsupervised}.
\begin{table*}[t]
	\begin{center}
		\centering
		\small
		\caption{Quantitative comparison with state-of-the-art methods. The best and second results are highlight in \textbf{bold} and \underline{underline}.  }\vspace{-1.1em}
		\label{tab:quantitative_result}
		\begin{tabular}{p{0.3cm}<{\centering}|p{0.8cm}<{\centering}|p{0.7cm}<{\centering}|p{0.7cm}<{\centering}|p{0.7cm}<{\centering}|p{0.7cm}<{\centering}|p{0.7cm}<{\centering}|p{0.7cm}<{\centering}|p{0.7cm}<{\centering}|p{0.7cm}<{\centering}|p{0.7cm}<{\centering}|p{0.7cm}<{\centering}|p{0.7cm}<{\centering}|p{0.7cm}<{\centering}|p{0.7cm}<{\centering}|p{0.7cm}<{\centering}|p{0.7cm}<{\centering}}
			\hline
			\multicolumn{2}{c|}{Dataset}&\multicolumn{4}{c|}{\cellcolor{green!20}MSIS}&\multicolumn{4}{c|}{\cellcolor{yellow!20}ChaMS-Real}&\multicolumn{7}{c}{\cellcolor{blue!20}ChaMS-Syn}\\\hline
			\multicolumn{2}{c|}{Metric}&\cellcolor{red!20}SF~$\uparrow$&\cellcolor{olive!20}SD~$\uparrow$&\cellcolor{teal!20}AG~$\uparrow$&\cellcolor{orange!20}BR~$\uparrow$&\cellcolor{red!20}SF~$\uparrow$&\cellcolor{olive!20}SD~$\uparrow$&\cellcolor{teal!20}AG~$\uparrow$&\cellcolor{orange!20}BR~$\uparrow$&\cellcolor{red!20}SF~$\uparrow$&\cellcolor{olive!20}SD~$\uparrow$&\cellcolor{teal!20}AG~$\uparrow$&\cellcolor{orange!20}BR~$\uparrow$&\cellcolor{purple!10}LPIPS~$\downarrow$&\cellcolor{lime!20}FID~$\downarrow$&\cellcolor{magenta!20}MSE~$\downarrow$\\\hline
			\multirow{5}{*}{\rotatebox{90}{FusionGAN }}&APAP&9.222&47.187&2.870&0.488&9.598&49.140&2.946&0.486    &11.522&48.104&4.223&0.478&0.309&69.833&9.818 \\
			&SPW &9.485&47.446&3.000&0.481&10.330&50.655&3.106&0.487    &9.024&41.197&4.728&0.487&0.486&151.363&10.185 \\
			&WPIS&9.494&45.155&3.173&\underline{0.496}&9.935&46.300&3.298&0.483     &11.300&47.040&4.417&0.485&0.393&81.886&9.903 \\
			&VFIS &9.832&42.777&3.604&0.488&10.301&44.759&3.697&0.487    &12.011&49.346&4.035&0.484&0.443&80.691&9.860 \\
			&RSFI &7.824&47.330&2.351&0.490&8.334&49.018&2.444&0.491    &11.161&48.940&3.746&0.495&0.408&67.107&9.896 \\\hline\hline
			
			\multirow{5}{*}{\rotatebox{90}{ GANMcC}}&APAP&9.228&48.659&2.853&0.490&9.710&50.900&2.898&0.489    &12.647&52.385&4.689&0.487&0.273&62.234&9.741 \\
			&SPW &9.175&48.808&2.837&0.487&9.740&50.401&2.875&0.491    &9.971&46.085&5.218&\textbf{0.498}&0.477&144.891&10.179 \\
			&WPIS &9.178&47.240&2.939&0.488&9.766&49.079&3.089&0.490    &12.376&51.359&4.838&0.487&0.385&74.836&9.887 \\
			&VFIS &9.485&43.188&3.477&0.489&9.934&44.633&3.561&0.489    &12.740&52.959&4.449&0.494&0.444&79.289&9.846 \\
			&RSFI &7.924&48.588&2.399&0.488&8.388&50.154&2.488&0.486    &11.756&52.677&4.088&0.486&0.409&60.636&9.871 \\\hline\hline

			\multirow{5}{*}{\rotatebox{90}{MFEIF }}&APAP&9.801&52.116&2.609&0.487&10.396&54.103&2.695&0.491    &16.279&62.692&4.775&0.490&0.137&27.419&5.918 \\
			&SPW &9.976&52.078&2.751&0.492&10.663&54.587&2.844&0.489    &10.948&44.577&\underline{5.273}&0.495&0.420&117.726&8.402 \\
			&WPIS &9.854&48.460&2.826&0.488&10.704&52.180&2.873&0.485    &16.392&62.672&4.878&0.490&0.293&39.412&7.548 \\
			&VFIS &9.548&44.200&3.240&0.487&10.115&45.424&3.355&0.491    &\underline{16.462}&\underline{65.555}&4.486&0.491&0.318&51.700&7.661 \\
			&RSFI &8.723&51.315&2.348&0.482&9.169&52.261&2.441&0.488     &15.673&64.347&4.216&0.493&0.352&38.787&8.080 \\\hline\hline

			\multirow{5}{*}{\rotatebox{90}{RFN }}&APAP&9.217&49.065&2.663&0.486&9.505&50.100&2.626&0.490    &12.662&55.140&4.337&0.488&0.191&52.365&9.990 \\
			&SPW &9.235&48.727&2.741&0.491&9.607&50.798&2.680&0.485    &9.703&48.910&4.719&0.494&0.429&136.680&10.241 \\
			&WPIS &9.361&47.695&2.824&0.486&9.608&49.486&2.756&0.488    &12.571&55.219&4.398&0.489&0.316&62.291&9.999 \\
			&VFIS &9.188&42.498&3.297&\textbf{0.497}&9.447&42.977&3.287&0.482    & 12.516&55.741&4.056&0.487&0.397&71.493&9.963\\
			&RSFI &8.296&47.451&2.504&0.489&8.640&47.768&2.543&0.486    &11.899&55.505&3.906&0.488&0.369&56.107&10.017 \\\hline\hline
			
			\multirow{5}{*}{\rotatebox{90}{ReCoNet }}&APAP&12.010&\underline{57.827}&3.410&0.485&11.779&\underline{57.119}&3.231&0.487    &15.974&62.277&4.766&0.492&\underline{0.134}&\textbf{27.174}&\underline{5.976} \\
			&SPW &13.004&55.174&3.462&0.490&11.742&57.027&3.320&0.485    &11.254&46.283&5.247&0.493&0.408&104.617&8.311 \\
			&WPIS &12.740&52.380&3.503&0.493&12.132&55.814&3.469&\underline{0.493}    &16.092&62.158&4.925&0.490&0.284&38.510&7.436 \\
			&VFIS &\underline{13.295}&47.977&\textbf{4.086}&0.493&\underline{12.410}&48.539&\textbf{4.058}&0.492    &15.868&64.280&4.474&0.492&0.337&62.081&7.815 \\
			&RSFI &10.315&52.496&2.997&0.486&10.312&53.371&3.043&0.486    &15.413&63.029&4.289&0.492&0.347&36.803&8.047 \\\hline\hline
			\multicolumn{2}{c|}{\textbf{Ours}}&\cellcolor{gray!40}\textbf{13.562}&\cellcolor{gray!40}\textbf{57.886}&\cellcolor{gray!40}\underline{3.628}&\cellcolor{gray!40}\underline{0.496}&\cellcolor{gray!40}\textbf{12.511}&\cellcolor{gray!40}\textbf{57.352}&\cellcolor{gray!40}\underline{3.883}&\cellcolor{gray!40}\textbf{0.494}    &\cellcolor{gray!40}\textbf{17.872}&\cellcolor{gray!40}\textbf{67.672}&\cellcolor{gray!40}\textbf{5.792}&\cellcolor{gray!40}\underline{0.496}&\cellcolor{gray!40}\textbf{0.131}&\cellcolor{gray!40}\underline{27.300}&\cellcolor{gray!40}\textbf{5.851} \\\hline
		\end{tabular}
	\end{center}
\end{table*}
\begin{figure*}[]
	\centering
	\setlength{\tabcolsep}{1pt}
	\begin{tabular}{cccccccccccc}	
		\includegraphics[width=0.08\textwidth,height=0.080\textheight]{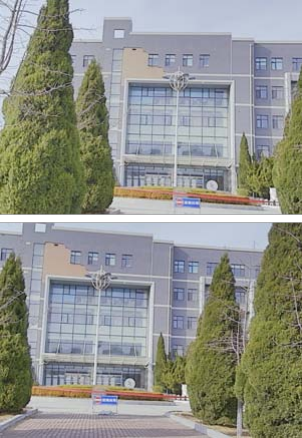}
		&\includegraphics[width=0.08\textwidth,height=0.080\textheight]{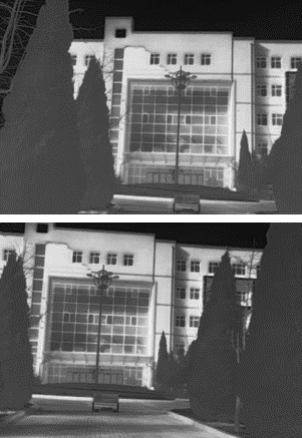}
		&\includegraphics[width=0.25\textwidth,height=0.080\textheight]{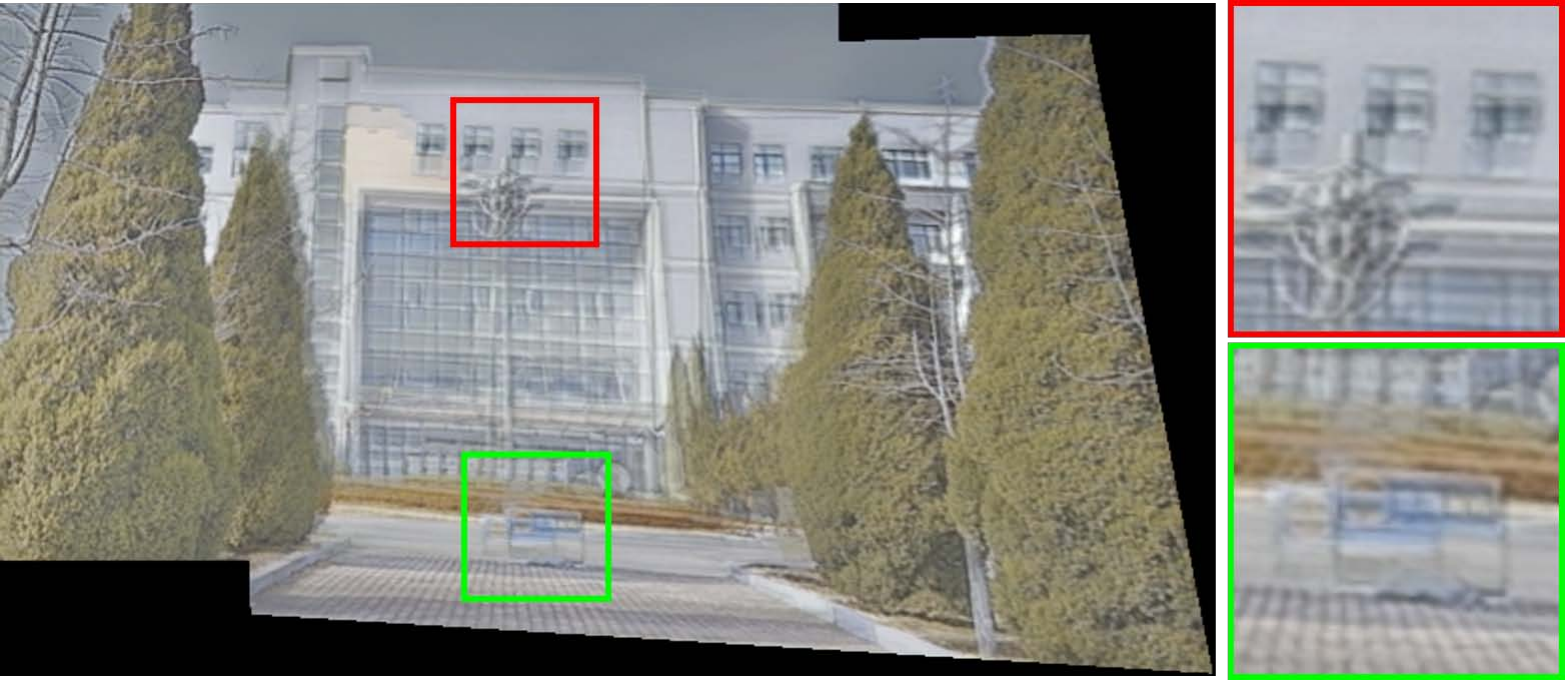}
		&\includegraphics[width=0.25\textwidth,height=0.080\textheight]{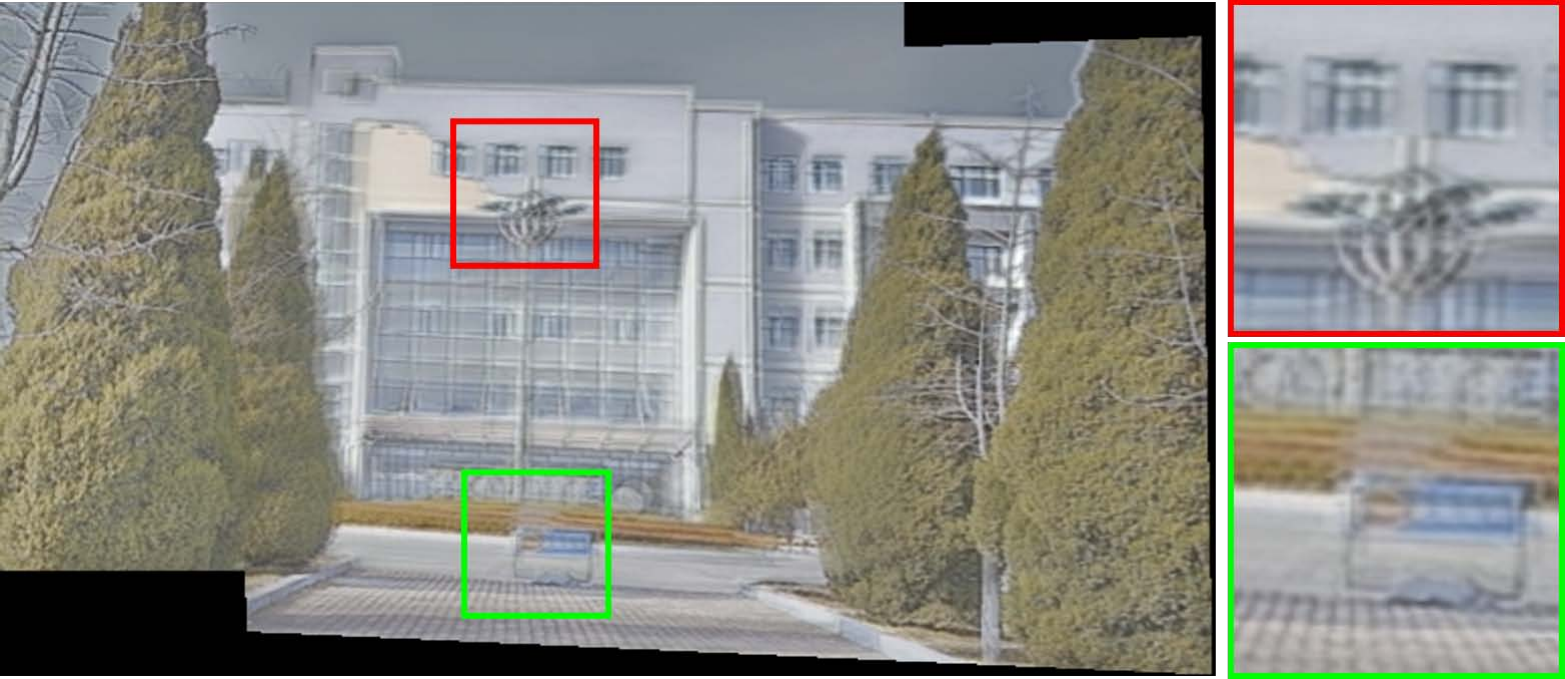}
		&\includegraphics[width=0.25\textwidth,height=0.080\textheight]{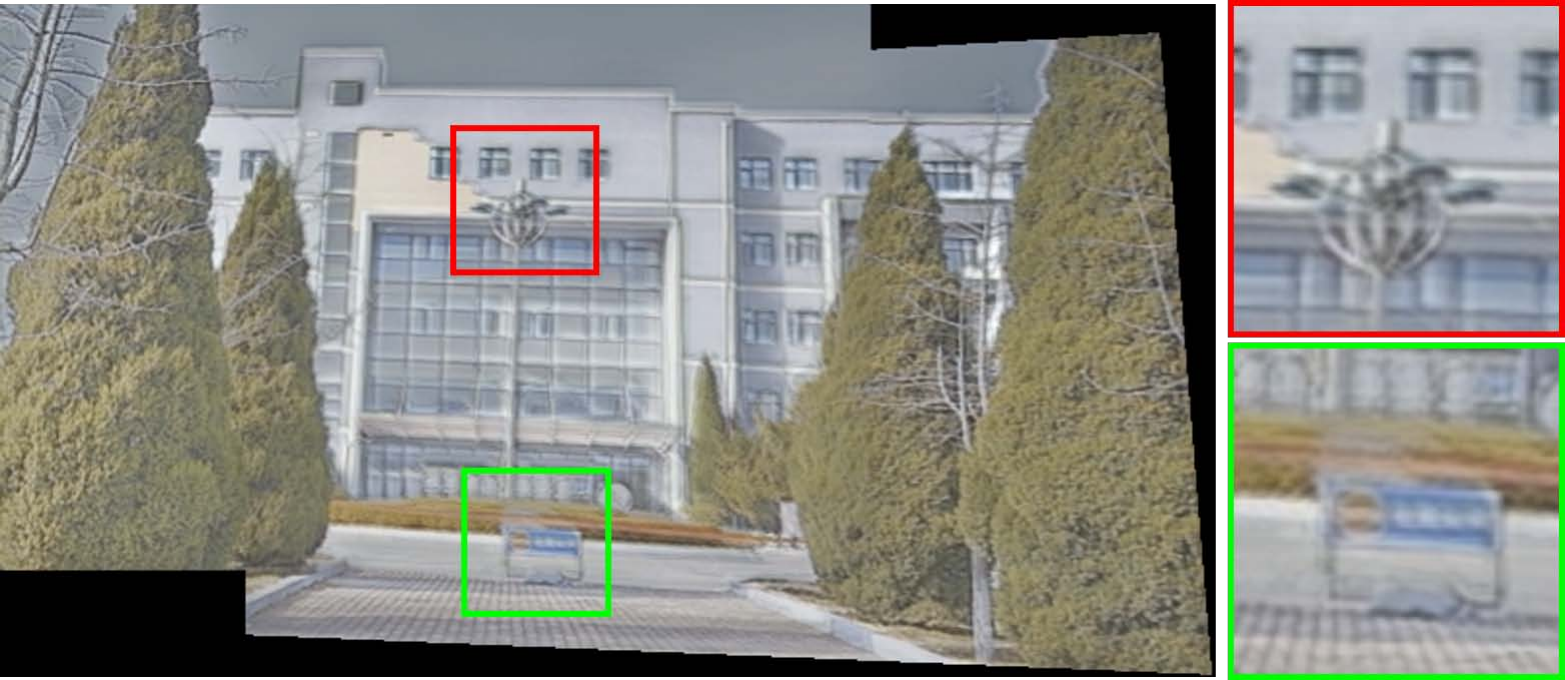}\\
		\includegraphics[width=0.08\textwidth,height=0.080\textheight]{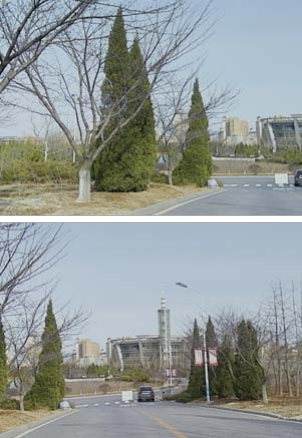}
		&\includegraphics[width=0.08\textwidth,height=0.080\textheight]{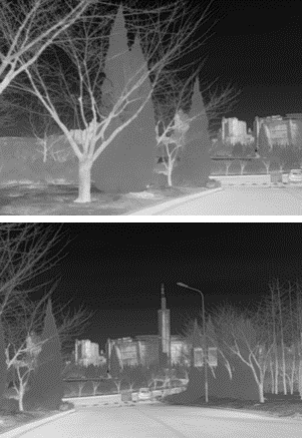}
		&\includegraphics[width=0.25\textwidth,height=0.080\textheight]{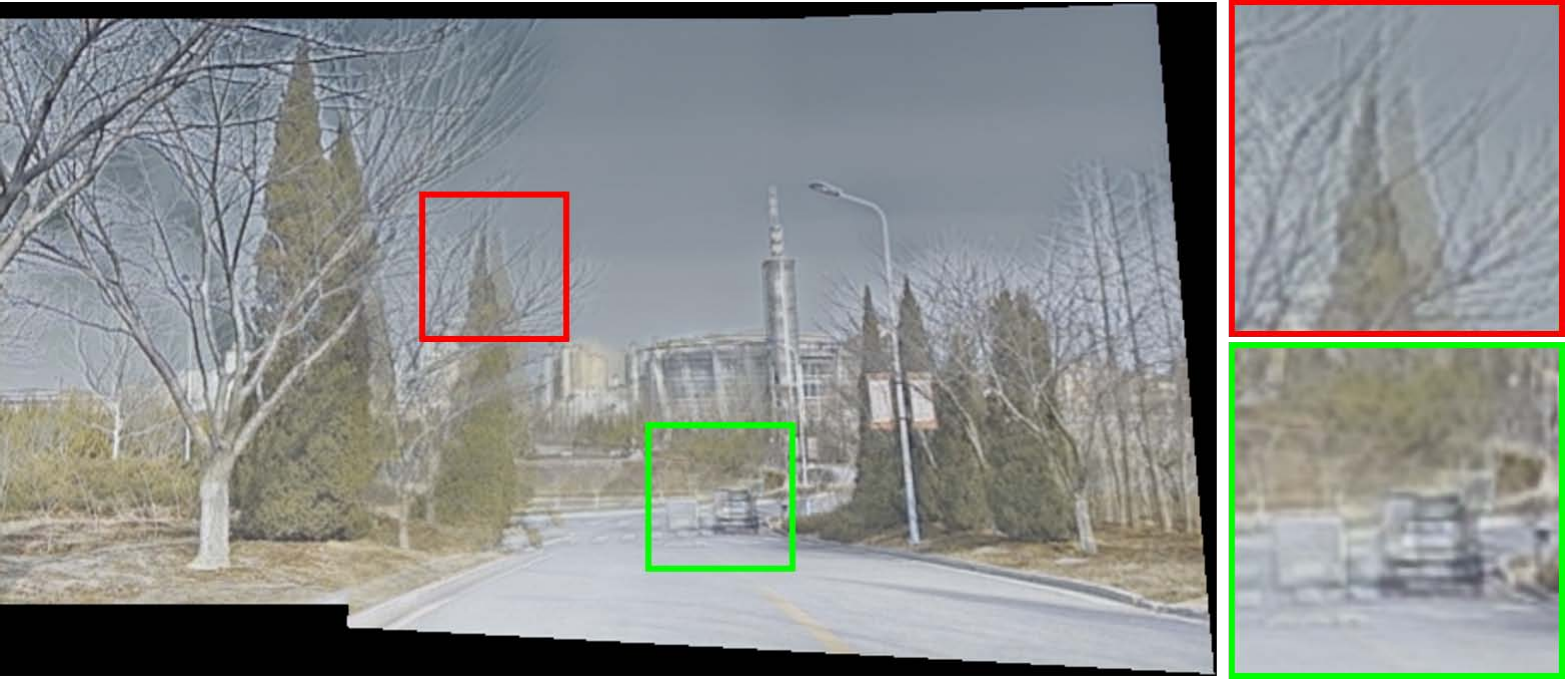}
		&\includegraphics[width=0.25\textwidth,height=0.080\textheight]{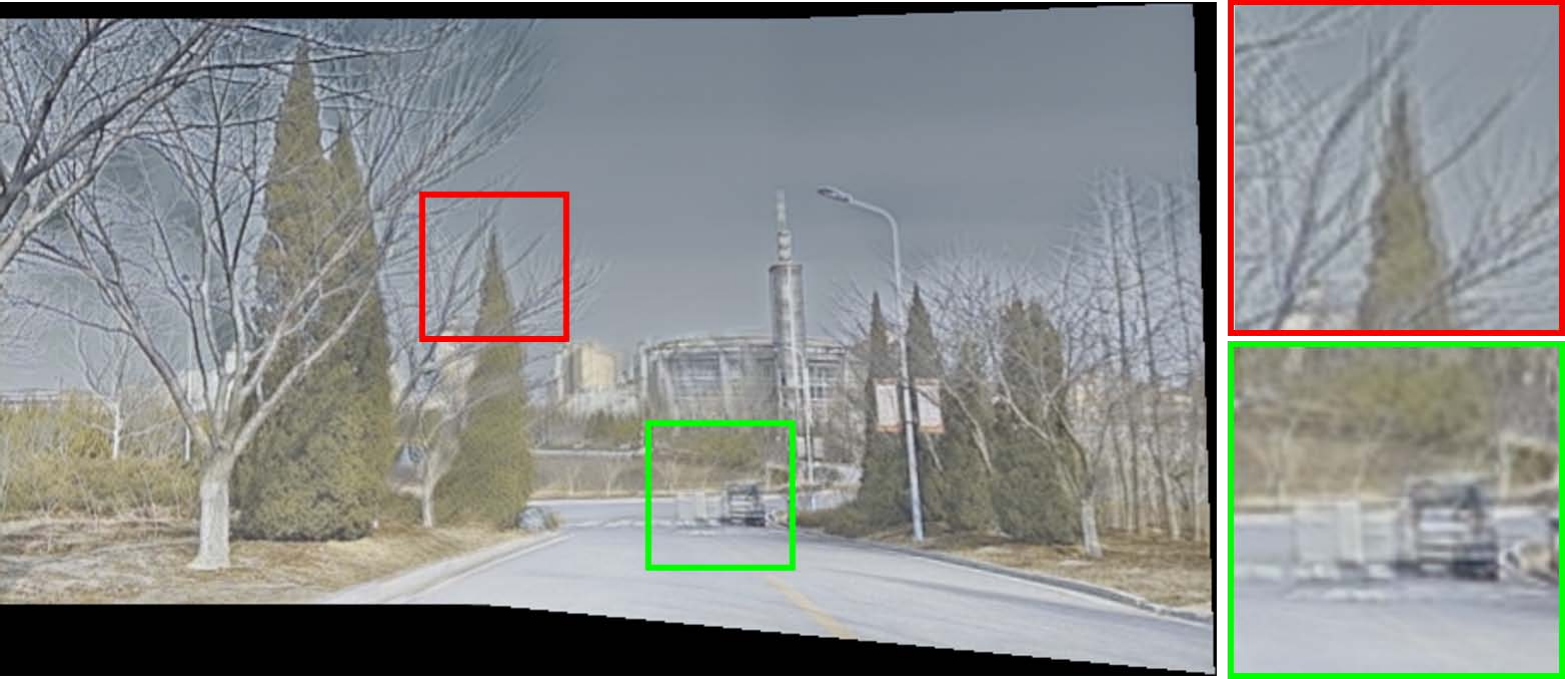}
		&\includegraphics[width=0.25\textwidth,height=0.080\textheight]{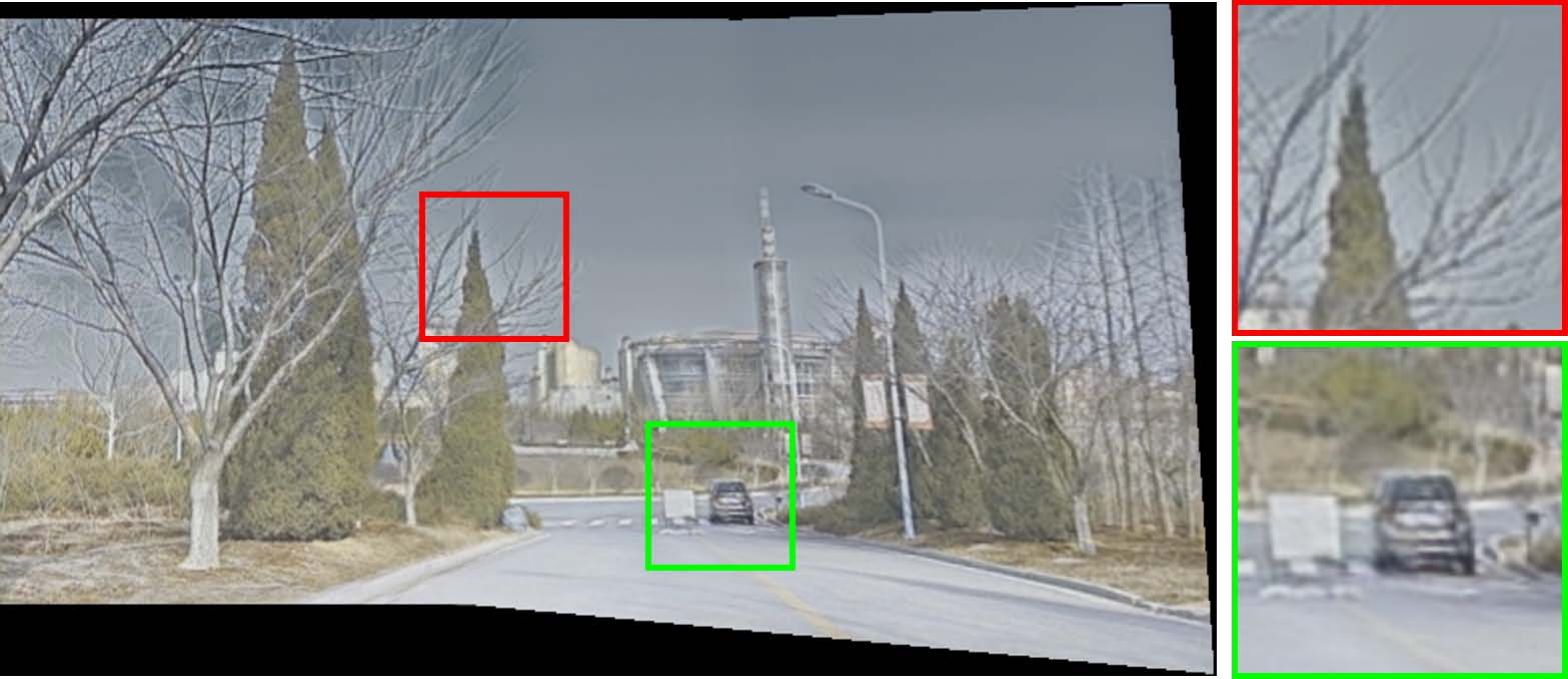}\\
		VIS&IR&GC&CCL&Ours\\
	\end{tabular}\vspace{-1.4em}
	\caption{ Ablation study on the proposed graph based relation reasoning with the conventional correlation based alignment strategy~(including GC, CCL). The proposed method achieves a more accurate alignment, free from the ghosting effect.}\vspace{-1em}
	\label{fig:ablation_cost}
\end{figure*}
 
\vspace{-0.5em}
\subsubsection{Results on MSIS Dateset}
The first row of Fig.~\ref{fig:vis_MSIS_challenge} presents a visual comparison of the results obtained from different stitching algorithms on MSIS, using the fusion process of ReCoNet. We observe that APAP exhibits a significant seam in the generated wide FOV scene, while both WPIS and VFIS methods suffer from alignment errors, resulting in a significant loss of scene content. SPW and RSFI display a certain degree of blurring in the overlapping regions. In contrast, the result of the proposed method demonstrates clear and accurate structures without introducing adverse interference.

Quantitative comparison is provided in Table.~\ref{tab:quantitative_result}, where Spatial Frequency~(SF)~\cite{eskicioglu1995image}, Standard Deviation~(SD)~\cite{rao1997fibre}, Average Gradient~(AG)~\cite{cui2015detail}, and Blind/Referenceless image spatial quality evaluator~(BR)~\cite{mittal2012no} are employed as metrics. The stitching performance positively correlates to the metric value. The proposed method performs the best in terms of SF and SD, while ReCoNet-VFIS and RFN-VFIS rank first in AG and BR, with the proposed method ranking a close second albeit with a slight disadvantage.
\vspace{-0.5em}
\subsubsection{Results on ChaMS Dataset}
Qualitative results on the ChaMS-Real dataset are illustrated in the second row of Fig.~\ref{fig:vis_MSIS_challenge}. Since this dataset focuses more on challenging large baseline and parallax scenes, the deep learning based VFIS and RSFI exhibit alignment failures, with the stitching results barely presenting the target scene content. Conventional methods such as APAP, SPW, and WPIS introduce noticeable stitching seam interference, resulting in unpleasant scene transitions. In contrast, the proposed method demonstrates a significant advantage. It is worth noting that the proposed method considers multi-spectral image information during the reconstruction process, with the introduction of infrared images leading to lower flamboyance in the generated results. Meanwhile, the competitive ReCoNet based strategies place more emphasis on preserving visible images, the scenes that are closer to those in visible images.

Fig.~\ref{fig:vis_syn} presents visual results obtained on the ChaMS-Syn dataset. The second example poses a significant challenge for the alignment across varying viewpoints due to its low quality and blurred scene. As a result, all competing methods display stitching failures, which manifest as distortion and content loss. In contrast, the proposed method more convincingly reconstructs the wide FOV scene.


Quantitative results are illustrated in Table.~\ref{tab:quantitative_result}. Since the ground truth of ChaMS-Syn is available, in addition to the unsupervised metrics, we also employed three supervised metrics, including Learned Perceptual Image Patch Similarity~(LPIPS)~\cite{zhang2018unreasonable}, Fréchet Inception Distance~(FID)~\cite{heusel2017gans}, and MSE, where the stitching performance is negatively correlated with these metric values. It can be observed that the proposed method achieves the best results in terms of LPIPS and MSE, and ranks second in the FID metric. The comprehensive analysis of the qualitative and quantitative results reveals the superiority of the proposed method over others.

\vspace{-.4em}
\subsection{Ablation Study}
We carried out an extensive ablation study and analysis to validate the effectiveness of the proposed method.
\begin{figure*}[t]
	\centering
	\setlength{\tabcolsep}{1pt}
	\begin{tabular}{cccccccccccc}	
		\includegraphics[width=1\textwidth, height=0.125\textheight]{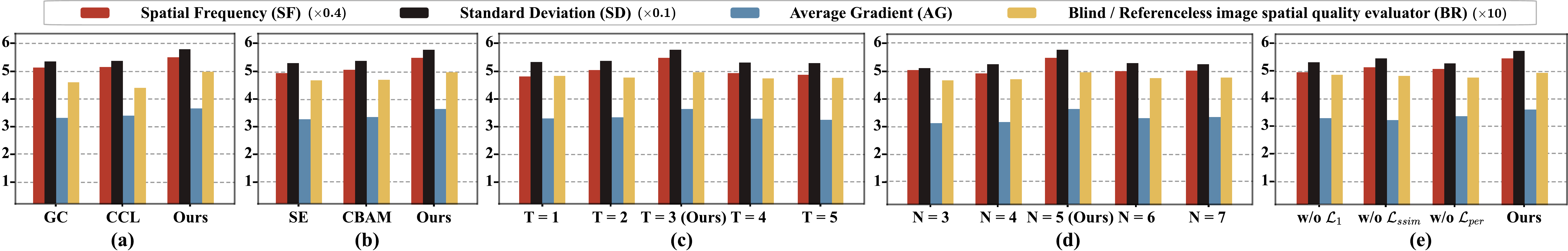}\\
	\end{tabular}\vspace{-0.45cm}
	\caption{The visualized objective comparison in ablation analysis. (a) graph based correlation reasoning, (b) the reconstruction module, (c) the number of progressive reasoning, (d) node number and (e) the loss function. }\vspace{-0.1cm}
	\label{fig:ablation}
\end{figure*}
\begin{figure*}[]
	\centering
	\setlength{\tabcolsep}{1pt}
	\begin{tabular}{cccccccccccc}	
		\includegraphics[width=0.08\textwidth,height=0.080\textheight]{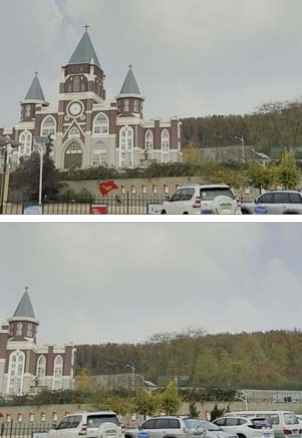}
		&\includegraphics[width=0.08\textwidth,height=0.080\textheight]{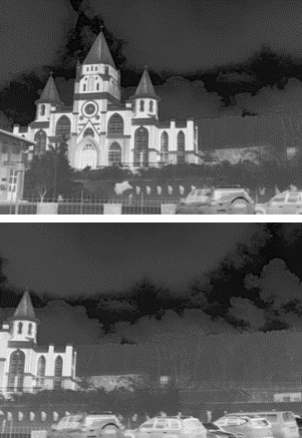}
		&\includegraphics[width=0.25\textwidth,height=0.080\textheight]{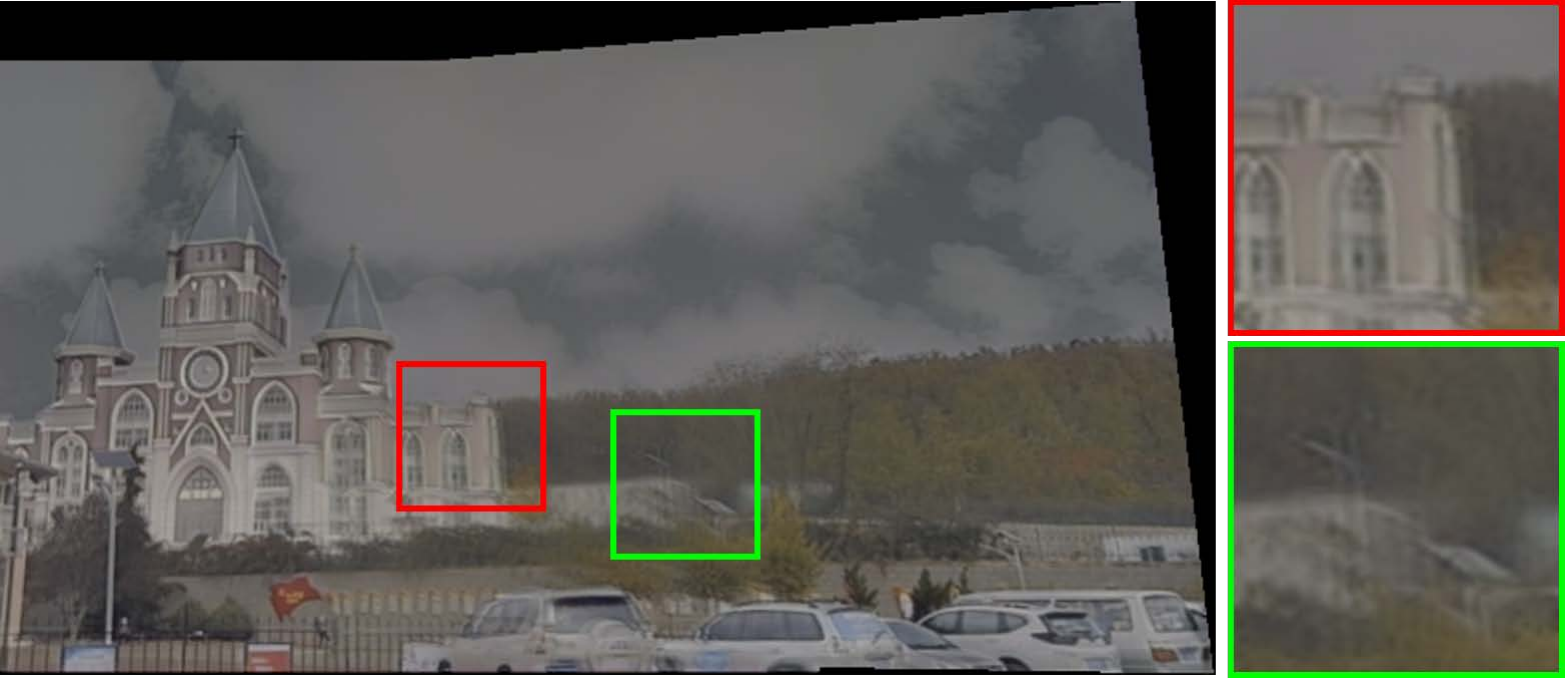}
		&\includegraphics[width=0.25\textwidth,height=0.080\textheight]{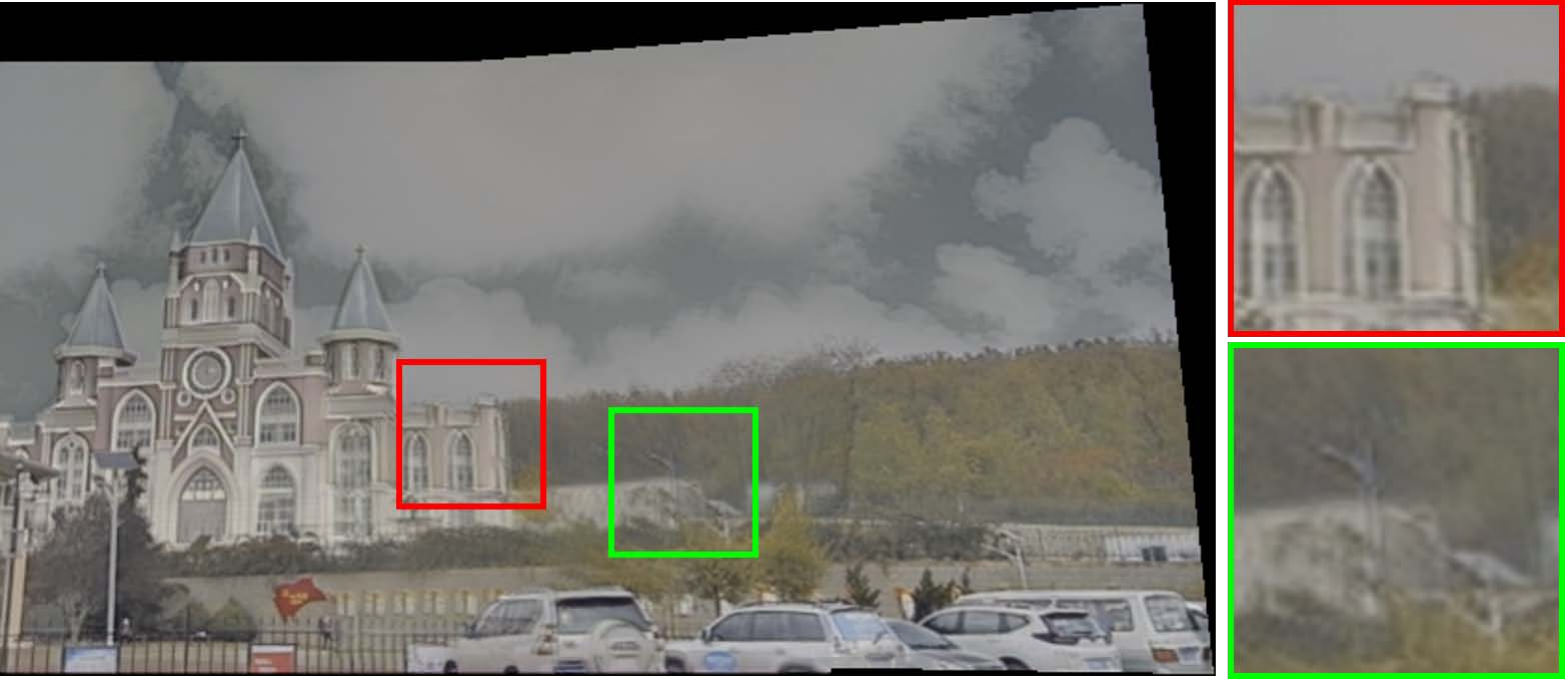}
		&\includegraphics[width=0.25\textwidth,height=0.080\textheight]{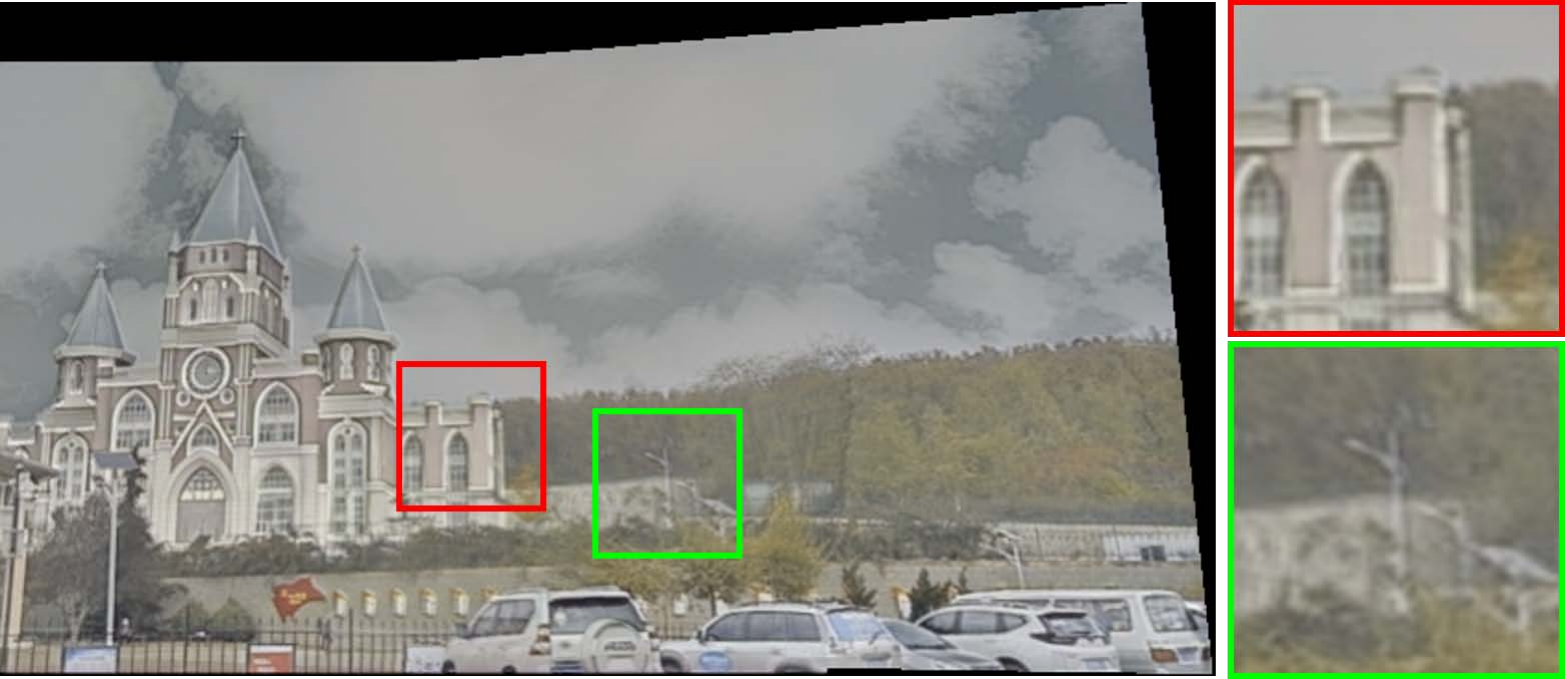}\\
		\includegraphics[width=0.08\textwidth,height=0.080\textheight]{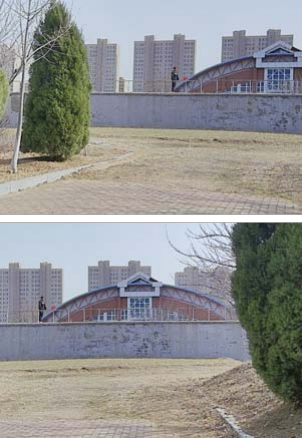}
		&\includegraphics[width=0.08\textwidth,height=0.080\textheight]{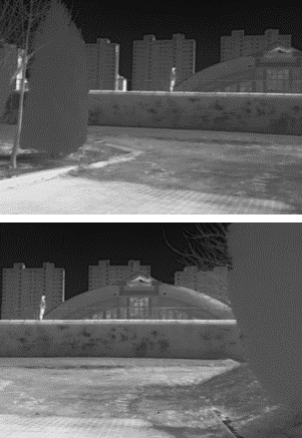}
		&\includegraphics[width=0.25\textwidth,height=0.080\textheight]{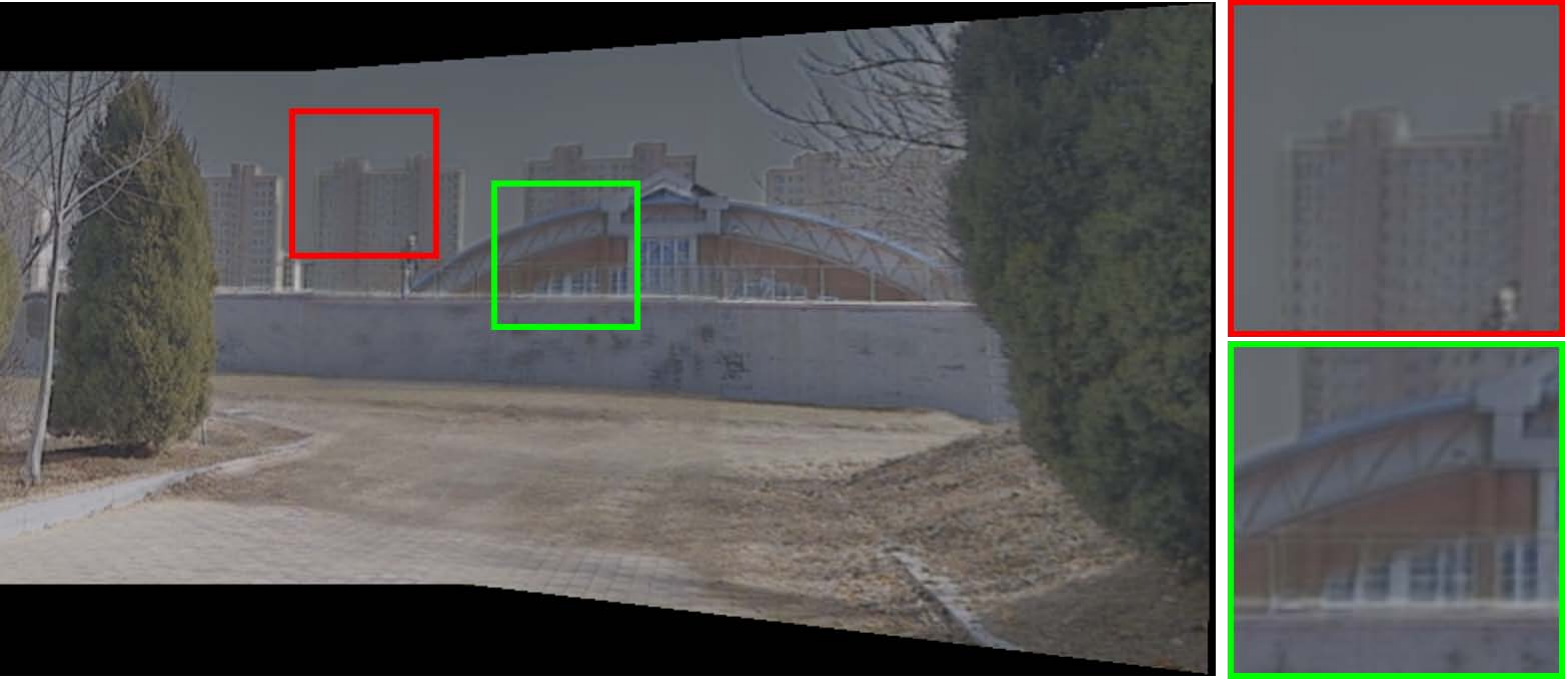}
		&\includegraphics[width=0.25\textwidth,height=0.080\textheight]{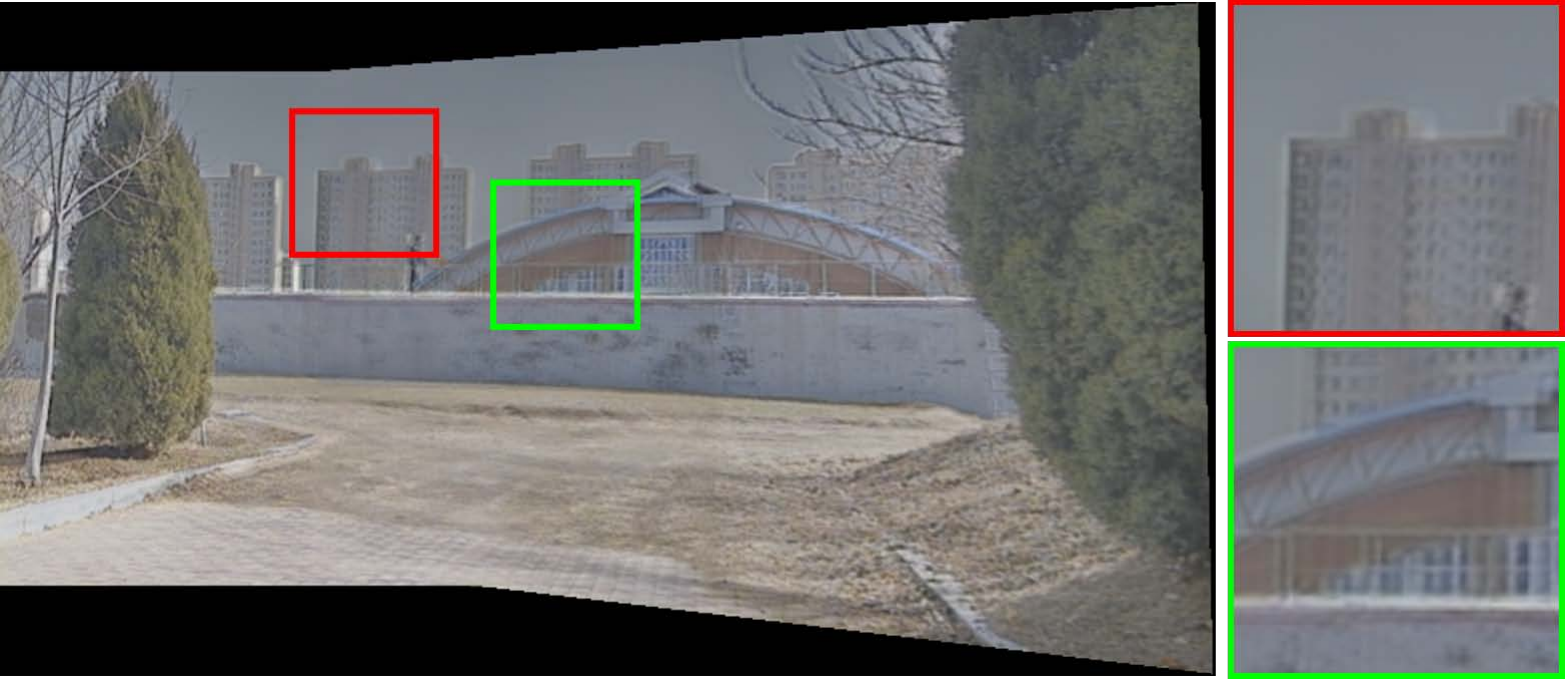}
		&\includegraphics[width=0.25\textwidth,height=0.080\textheight]{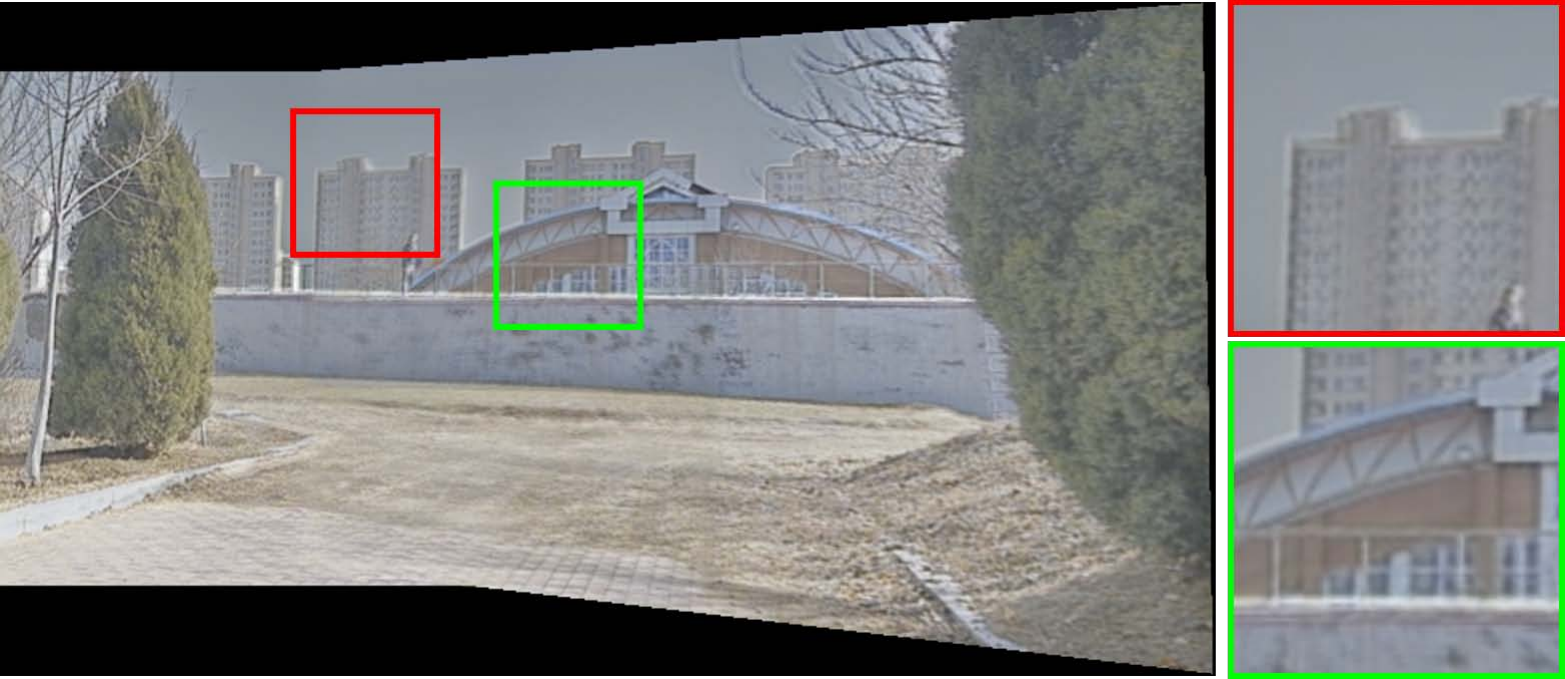}\\
		VIS&IR&SE&CBAM&Ours\\
	\end{tabular}\vspace{-0.45cm}
	\caption{Ablation study on the proposed reconstruction module with the channel attention based reconstruction.}
	\vspace{-0.3cm}
	\label{fig:ablation_attention}
\end{figure*}
\vspace{-.3em}
\subsubsection{Graph Reasoning vs. Correlation Matching}
The alignment across multi-spectral multi-view scenes is carried out with the proposed inter- and intra-correlation based graph reasoning. To validate its effectiveness, we conducted a comparison with the conventional correlation calculation based strategies, including global correlation~(GC)~\cite{Sun_2018_CVPR} and contextual correlation~(CCL)~\cite{nie2021depth}. Visual results are presented in Fig.~\ref{fig:ablation_cost}. While GC entails pixel-wise correlation, it ignores long-range coherence, resulting in diminished alignment effectiveness and prominent ghosting artifacts in the stitched results. The CCL strategy also encounters problems with ghosting effects. Conversely, our method facilitates the relation reasoning among nodes from both  different view positions and the same position, offering a marked advantage in cross-view alignment. Quantitative results are illustrated in fig.~\ref{fig:ablation}~(a), in which the proposed method achieves the best in all metrics. 
\begin{figure}[]
	\begin{minipage}{0.116\textwidth}
		\centerline{\includegraphics[width=1\textwidth,height=0.06\textheight]{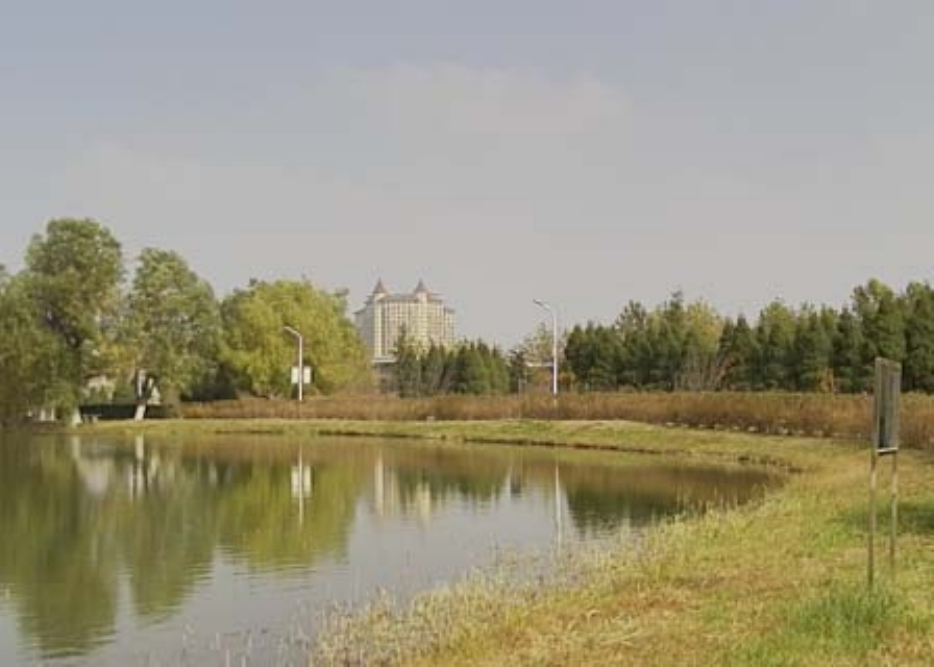}}\vspace{-.3em}
		\centerline{VIS 1}
	\end{minipage}
	\hfill
	\begin{minipage}{0.116\textwidth}
		\centerline{\includegraphics[width=1\textwidth,height=0.06\textheight]{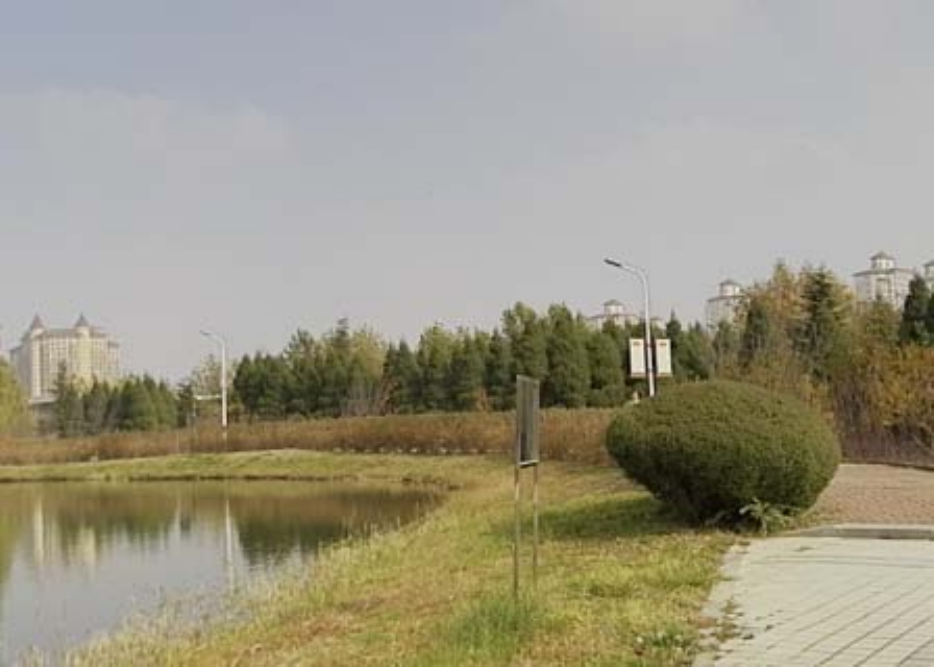}}\vspace{-.3em}
		\centerline{VIS 2}
	\end{minipage}
	\hfill
	\begin{minipage}{0.116\textwidth}
		\centerline{\includegraphics[width=1\textwidth,height=0.06\textheight]{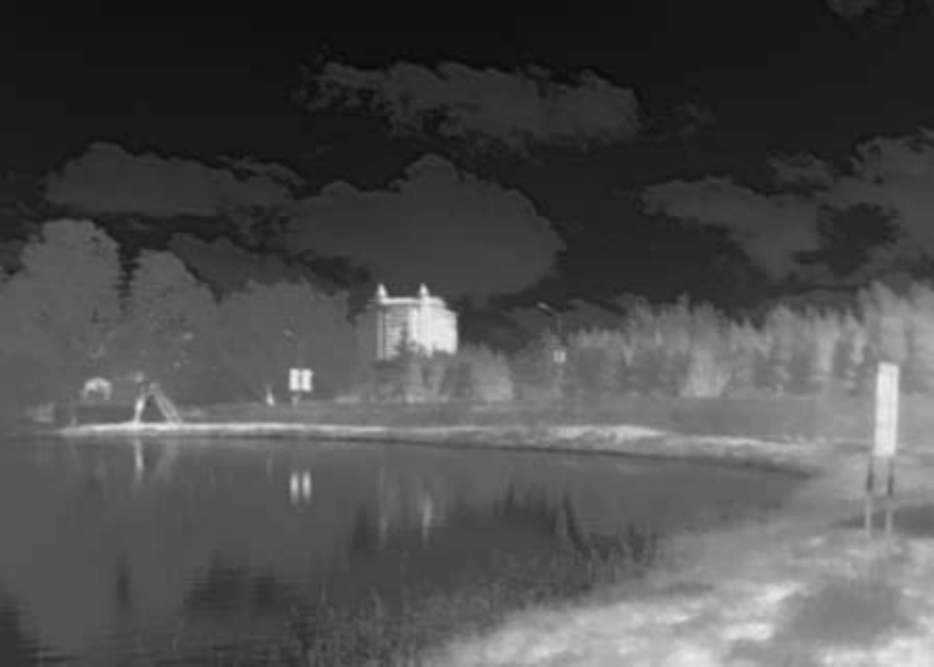}}\vspace{-.3em}
		\centerline{IR 1}
	\end{minipage}
	\hfill
	\begin{minipage}{0.116\textwidth}
		\centerline{\includegraphics[width=1\textwidth,height=0.06\textheight]{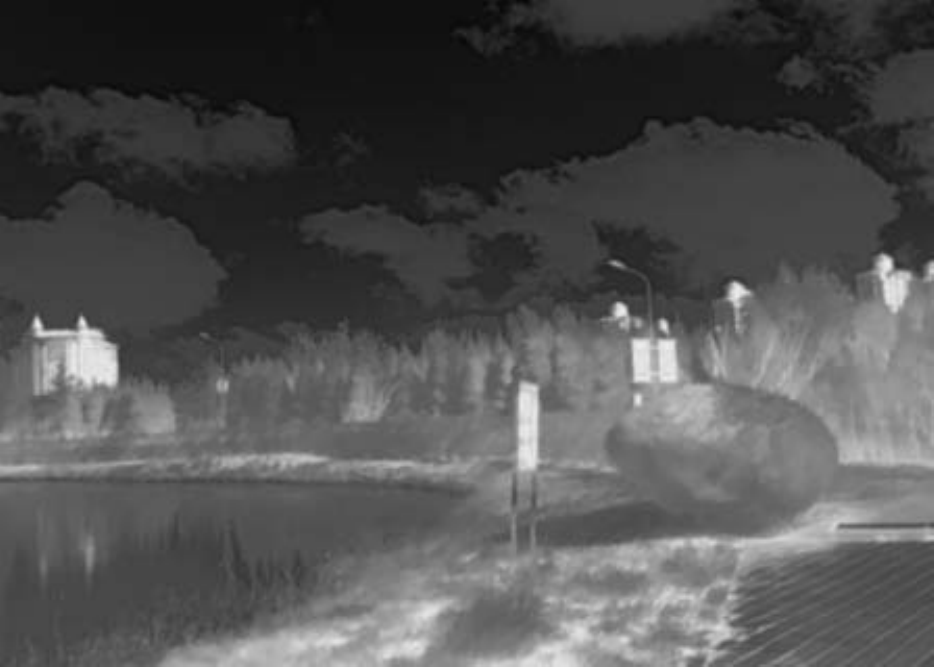}}\vspace{-.3em}
		\centerline{IR 2}
	\end{minipage}
	\vfill
	\begin{minipage}{0.236\textwidth}
		\centerline{\includegraphics[width=1\textwidth,height=0.06\textheight]{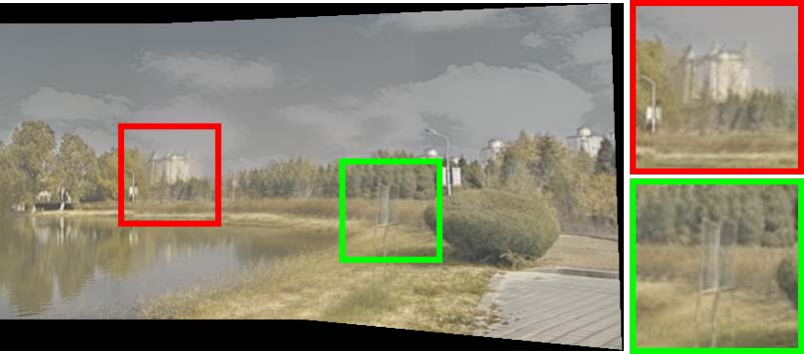}}\vspace{-.3em}
		\centerline{T = 2}
	\end{minipage}
	\hfill
	\begin{minipage}{0.236\textwidth}
		\centerline{\includegraphics[width=1\textwidth,height=0.06\textheight]{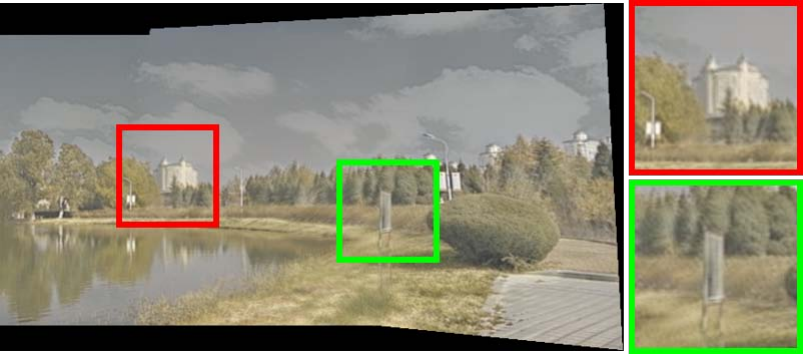}}\vspace{-.3em}
		\centerline{T = 3 (Ours)}
	\end{minipage}
	\vfill
	\begin{minipage}{0.236\textwidth}
		\centerline{\includegraphics[width=1\textwidth]{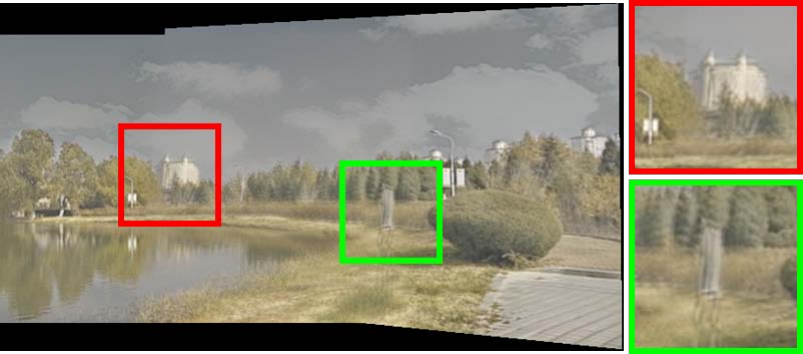}}\vspace{-.3em}
		\centerline{T = 4}
	\end{minipage}
	\hfill
	\begin{minipage}{0.236\textwidth}
		\centerline{\includegraphics[width=1\textwidth]{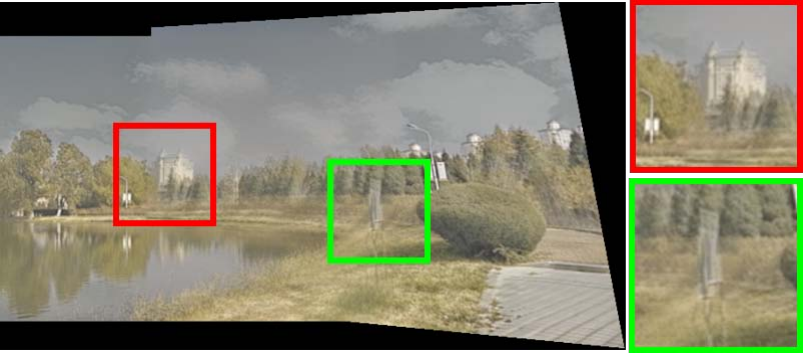}}\vspace{-.3em}
		\centerline{T = 5}
	\end{minipage}
	\vfill
	\vspace{-0.30cm}
	\caption{Ablation study on the progressive number.}
	\vspace{-0.7cm}
	\label{fig:ablation_T}
\end{figure}

\vspace{-.5em}
\subsubsection{Graph Reasoning vs. Channel Attention}
The reconstruction phase not only achieves the integration of multi-view scenes but also the fusion of multi-spectral images. To confirm the effectiveness of spatial and channel graph based reconstruction, we conducted a comparison with conventional attention based strategies, including channel-wise attention block~(SE)~\cite{hu2018squeeze} and convolutional block attention module~(CBAM)~\cite{woo2018cbam}. The visual comparisons are presented in Fig.~\ref{fig:ablation_attention}. Obviously, the results of SE and CBAM display ghosting interference in the overlapping regions between the two viewpoints. Additionally, the pixel luminance of the stitching results, which take into account both infrared and visible images, appears to have relatively low intensity. Our method produces a credible and accurate combination for the reconstruction of common regions with plausible pixel intensity. The corresponding objective comparison is visualized in Fig.~\ref{fig:ablation}~(b). The consistency between visual and quantitative performances validates the effectiveness of the proposed spatial and channel graph based reconstruction.

\vspace{-.3em}
\subsubsection{Number of Progressive Reasoning}
The inter and intra correlation based graph reasoning employs a progressive mechanism, in which the prior reasoning is used for the initialization of subsequent nodes. To explore the optimal structure, we compared the performance with different numbers of progressive reasoning~(denoted as T). The visual results are illustrated in Fig.~\ref{fig:ablation_T}, and the quantitative comparisons are presented in Fig.~\ref{fig:ablation}~(c). We found that the performance is optimal when T is set to~$3$. Consequently, we adopt a setting with three progressive reasoning.

\vspace{-.3em}
\subsubsection{Node Number}
We conducted experiments to investigate the impact of varying the number of nodes within the graph structure on the performance of our method.  The results shown in Fig.~\ref{fig:ablation}~(d) indicated that utilizing a graph structure comprising five nodes yielded optimal performance. Consequently, we implemented the proposed method with five nodes throughout the experiments.

\vspace{-.3em}
\subsubsection{Loss Validation}
Fig.~\ref{fig:ablation}~(e) illustrates the ablation study on the loss function. It can be observed that each loss term contributes to an improvement in the results. Therefore, the optimal multi-spectral image stitching performance is achieved when these loss functions are combined.
%

\section{Conclusion}
This paper proposed a spatial graph reasoning based multi-spectral image stitching method to generate wide FOV images with complementary and comprehensive information. We first investigate the inter- and intra-correlation within the embedded graph structure to facilitate multi-spectral relation reasoning, enhancing the cross-view alignment. During the reconstruction phase, we leverage long-range coherence to improve context perception along spatial and channel dimensions, promoting the integration of multi-view scenes and the fusion of multi-spectral images. These two graph based structures effectively manage the distillation and enrichment of multi-spectral feature representations. Extensive experiments conducted on MSIS and ChaMS datasets substantiate the superior performance of our method compared to alternative strategies.

\begin{acks}
	This work is partially supported by the National Key R\&D Program of China (No. 2022YFA1004101), the National Natural Science Foundation of China (No. U22B2052).
\end{acks}
\balance
\newpage
\bibliographystyle{ACM-Reference-Format}
\bibliography{ACMMM22_short}

\appendix

\end{document}